\documentclass[a4paper,fleqn]{cas-dc}

\usepackage[numbers]{natbib}
\newcommand{\bv}[1]{\textcolor{blue}{#1}} 
\newcommand{\rb}[1]{\textbf{\textcolor{red}{#1}}} 

\def\tsc#1{\csdef{#1}{\textsc{\lowercase{#1}}\xspace}}
\tsc{MRSI}
\tsc{CNN}
\tsc{GNN}

\begin{document}
\let\WriteBookmarks\relax
\def\floatpagepagefraction{1}
\def\textpagefraction{.001}

\makeatletter
\setlength{\@fptop}{0pt}        
\setlength{\@fpsep}{8pt}        
\setlength{\@fpbot}{0pt plus 1fil}  
\setlength{\@dblfptop}{0pt}     
\setlength{\@dblfpsep}{8pt}
\setlength{\@dblfpbot}{0pt plus 1fil}
\makeatother

\shorttitle{SOMA-1M: A Large-Scale SAR-Optical Multi-resolution Alignment Dataset for Multi-Task Remote Sensing}
\shortauthors{P. Wu et~al.}

\title [mode = title]{SOMA-1M: A Large-Scale SAR-Optical Multi-resolution Alignment Dataset for Multi-Task Remote Sensing}

\author[1]{Peihao Wu}
\fnmark[1]

\author[1,2,3]{Yongxiang Yao}[orcid=0000-0001-5492-0564]
\cormark[1] 

\author[1,2,3]{Yi Wan}

\author[1]{Wenfei Zhang}

\author[1]{Ruipeng Zhao}

\author[1]{Jiayuan Li}

\author[1,2,3]{Yongjun Zhang}[orcid=0000-0001-9845-4251]
\cormark[1] 

\affiliation[1]{organization={School of Remote Sensing Information Engineering, Wuhan University}, city={Wuhan}, postcode={430079}, country={China}}

\affiliation[2]{organization={Hubei LuoJia Laboratory}, 
                city={Wuhan}, 
                postcode={430079}, 
                country={China}}

\affiliation[3]{organization={Technology Innovation Center for Collaborative Applications of Natural Resources Data in GBA, Ministry of Natural Resources}, 
                city={Guangzhou},
                postcode={510075},
                country={China}}

\cortext[cor1]{Corresponding authors: Yongxiang Yao. E-mail address: yaoyongxiang@whu.edu.cn; Yongjun Zhang. E-mail address: zhangyj@whu.edu.cn.}

\fntext[fn1]{This is the first author.}

\begin{abstract}
Synthetic Aperture Radar (SAR) and optical imagery provide complementary strengths that constitute the critical foundation for transcending single-modality constraints and facilitating cross-modal collaborative processing and intelligent interpretation. However, existing benchmark datasets often suffer from limitations such as single spatial resolution, insufficient data scale, and low alignment accuracy, making them inadequate for supporting the training and generalization of multi-scale foundation models. To address these challenges, we introduce SOMA-1M (SAR-Optical Multi-resolution Alignment), a pixel-level precisely aligned dataset containing over 1.3 million pairs of georeferenced images with a specification of 512 × 512 pixels. This dataset integrates imagery from Sentinel-1, PIESAT-1, Capella Space, and Google Earth, achieving global multi-scale coverage from 0.5 m to 10 m. It encompasses 12 typical land cover categories, effectively ensuring scene diversity and complexity. To address multimodal projection deformation and massive data registration, we designed a rigorous coarse-to-fine image matching framework ensuring pixel-level alignment. Based on this dataset, we established comprehensive evaluation benchmarks for four hierarchical vision tasks, including image matching, image fusion, SAR-assisted cloud removal, and cross-modal translation, involving over 30 mainstream algorithms. Experimental results demonstrate that supervised training on SOMA-1M significantly enhances performance across all tasks. Notably, multimodal remote sensing image (MRSI) matching performance achieves current state-of-the-art (SOTA) levels. SOMA-1M serves as a foundational resource for robust multimodal algorithms and remote sensing foundation models. The dataset will be released publicly at: https://github.com/PeihaoWu/SOMA-1M.
\end{abstract}


\maketitle

\section{Introduction}\label{sec:introduction}

With the rapid advancement of remote sensing big data and artificial intelligence, Earth observation missions are transitioning from single-payload monitoring to multi-source collaborative intelligent interpretation \cite{zhu2017deep}. Optical remote sensing imagery, with its intuitive semantic textures and spectral features, has become the dominant data source for land cover classification and change detection; however, its imaging quality is severely constrained by atmospheric scattering, cloud occlusion, and lighting conditions, making it difficult to ensure continuity of monitoring \cite{wang2025mt_gan}. In contrast, Synthetic Aperture Radar (SAR) possesses all-weather imaging capabilities capable of penetrating clouds and rain, and offers unique advantages in sensing the geometric configuration and electromagnetic properties of terrestrial objects \cite{moreira2013tutorial}. This multimodal collaborative processing paradigm, which integrates the complementary strengths of multimodal imagery, has become a pivotal pathway for transcending the limitations of single modalities and achieving advanced cross-modal collaborative processing and intelligent interpretation \cite{schmitt2016data, ye20253mos}. From low-level geometric matching \cite{zhang2025soft, zhang2023histogram, zheng2025msg, zhang2025glift} to high-level cross-modal generation \cite{wang2025mt_gan, guo2024learning}, multimodal collaborative analysis demonstrates significant application potential; however, its performance upper bound is increasingly constrained by multiple factors regarding training data, including scale, precision, and diversity.

Although deep learning-based multimodal remote sensing research has achieved significant progress, enhancing model robustness and generalization in complex scenes remains a formidable challenge. The core bottleneck lies in the lack of large-scale, cross-resolution benchmark datasets with pixel-level alignment precision. While existing public datasets (e.g., SEN12MS \cite{schmitt2019sen12ms}, WHU-OPT-SAR \cite{li2022mcanet}) have provided initial momentum, they suffer from three notable limitations: (1) Single and coarse spatial resolution: Most datasets are predominantly constructed based on Sentinel-1/2, with resolutions restricted to the 10-m level. They lack sub-meter high-resolution samples, making them inadequate for supporting research on high-precision applications such as refined urban management. (2) Insufficient data scale: The sample size of existing datasets is typically on the order of tens of thousands of pairs, which fails to meet the training demands of large-parameter deep learning models regarding data diversity and generalization capabilities. (3) Inadequate alignment accuracy: This is the most critical issue. Due to significant differences in geometric imaging mechanisms between SAR and optical sensors, significant pixel deviations persist in areas with undulating terrain even after geocoding \cite{hughes2020deep}. Existing large-scale datasets often lack refined registration processing, resulting in local misalignment between image pairs. Such noisy data severely impedes model training for pixel-level tasks (e.g., image fusion and image translation).

To surmount these limitations, we propose SOMA-1M (SAR-Optical Multi-resolution Alignment), a large-scale multimodal remote sensing alignment dataset oriented towards multi-level tasks. SOMA-1M is currently the large-scale aligned dataset with the richest resolution diversity in the remote sensing field. The dataset comprises over 1.3 million pairs of georeferenced aligned images, spanning three typical resolution levels—0.5 m, 3 m, and 10 m. It integrates imagery from Sentinel-1, PIESAT-1, Capella Space, and Google Earth, achieving a multi-scale balance from global wide-area coverage to local fine-grained sampling. Addressing the nonlinear geometric distortions between multimodal images, we designed a rigorous coarse-to-fine automated registration framework. This framework utilizes a cascaded fine-tuning strategy to overcome nonlinear geometric distortions, ensuring pixel-level precise alignment. SOMA-1M aims to resolve the generalization challenges of multimodal remote sensing processing algorithms across scales and scenes.

\begin{figure*}[!t] 
    \centering
    \includegraphics[width=1.00\textwidth]{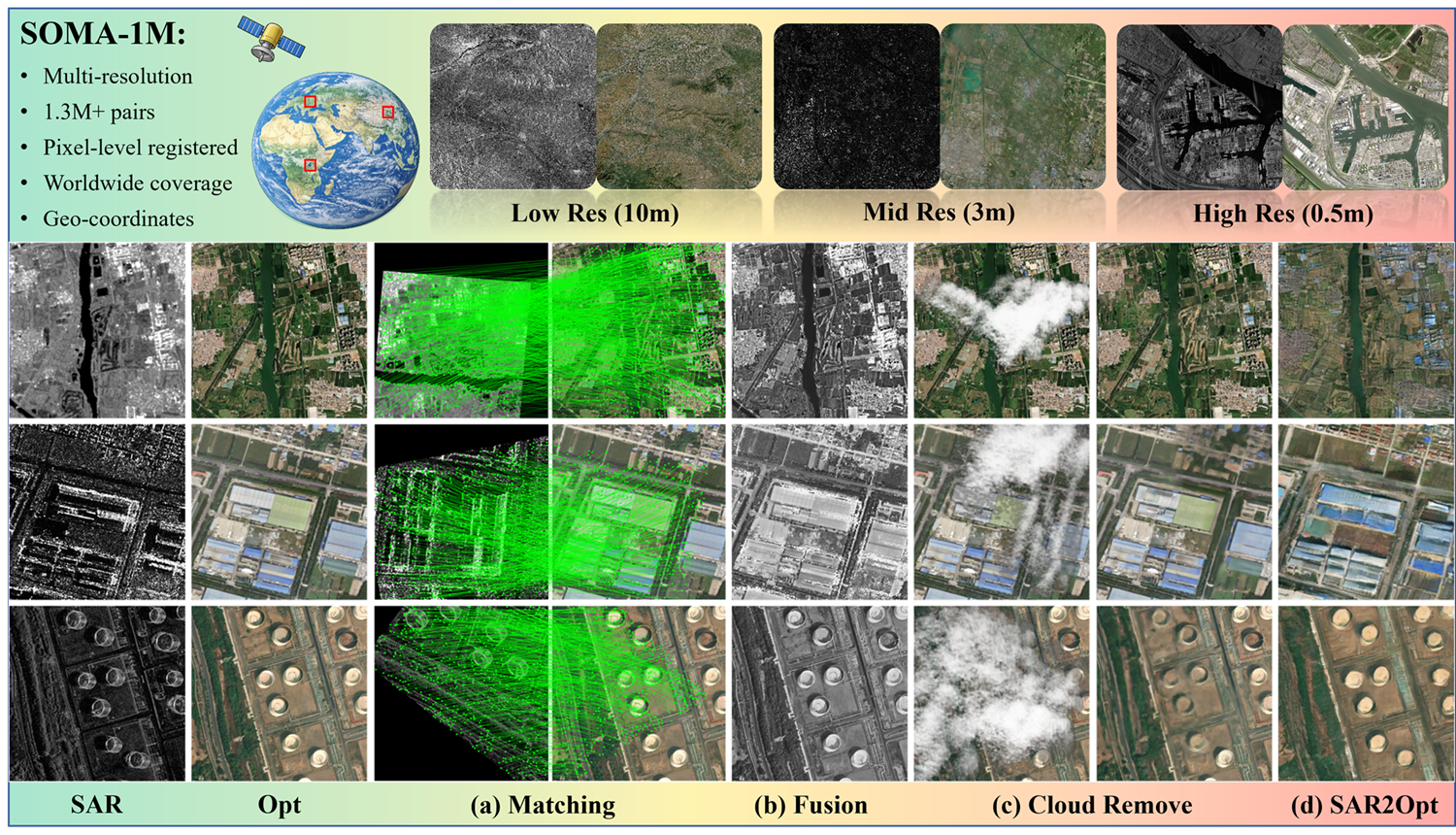}
    \caption{Overview of the SOMA-1M dataset and examples of its multi-task applications. The two leftmost columns display the original SAR and optical input images. The remaining columns illustrate representative results generated by models trained on this dataset: (a) Image Matching; (b) Image Fusion; (c) SAR-Assisted Cloud Removal; and (d) SAR-to-Optical Translation.}
    \label{fig:Overview of the SOMA-1M}
\end{figure*}

To validate the universality of SOMA-1M, this study established benchmarks encompassing four tasks: image matching, image fusion, SAR-assisted cloud removal, and SAR-to-optical translation. Qualitative results of the SOMA-1M dataset on these four benchmark tasks are illustrated in Fig. \ref{fig:Overview of the SOMA-1M}. Experiments demonstrate that SOMA-1M significantly activates the model's capability to extract cross-modal features. Furthermore, we delved into the domain gap introduced by resolution spans, revealing the limitations of models in parsing microscopic scattering mechanisms. The main contributions of this paper are summarized as follows:

\textbullet\ Proposed the SOMA-1M Dataset: By offering a million-scale data volume, a resolution span from 0.5-m to 10-m levels, and 12 categories of typical land cover, this dataset significantly expands the diversity of multimodal remote sensing data.

\textbullet\ Automated Annotation Method: We proposed a robust automated alignment processing framework for multimodal images, solving the challenge of high-precision SAR-optical registration under multi-resolution conditions and providing high-quality ground truth data for the community.

\textbullet\ Multi-level Benchmarking: Through exhaustive experiments on four representative vision tasks, we demonstrated the critical value of high-quality aligned samples in enhancing model generalization and robustness, and analyzed the sensitivity of different tasks to resolution.

\section{Related work}

\begin{table*}[t]
\centering
\caption{Comprehensive comparison of SOMA-1M with existing mainstream multimodal remote sensing datasets.}
\label{tab:soma_comparison}
\resizebox{\textwidth}{!}{
\begin{tabular}{l c l c c c c}
\toprule
\multicolumn{1}{c}{Datasets} & Year & \multicolumn{1}{c}{Modality \& GSD} & \#Pairs & Image Size (px) & Alignment & Multi-Res \\
\midrule
SEN12MS \cite{schmitt2019sen12ms} & 2019 & \{SAR, Opt\}(10m), LandCover(500m) & 180K & 256$\times$256 & Geographical-level & $\times$ \\
SEN12MS-CR \cite{ebel2020multisensor} & 2020 & \{Opt, SAR\}(10m), Cloud & 122K & 256$\times$256 & Geographical-level & $\times$ \\
So2Sat LCZ42 \cite{zhu2020so2sat} & 2020 & \{SAR, Opt\}(10m) & 400K & 32$\times$32 & Pixel-level & $\times$ \\
SpaceNet 6 \cite{shermeyer2020spacenet} & 2020 & \{SAR, Opt\}(0.5m) & 20K & 900$\times$900 & Pixel-level & $\times$ \\
OSdataset \cite{xiang2020automatic} & 2020 & \{SAR, Opt\}(1m) & 13K & \makecell[c]{512$\times$512 (10692) \\ 256$\times$256 (2673)} & Pixel-level & $\times$ \\
WHU-OPT-SAR \cite{li2022mcanet} & 2021 & \{SAR, Opt\}(5m) + Masks & 29K & 256$\times$256 & Pixel-level & $\times$ \\
QXS-SAROPT \cite{huang2021qxs} & 2021 & \{SAR, Opt\}(1m) & 20K & 256$\times$256 & Pixel-level & $\times$ \\
SSL4EO-S12 \cite{wang2023ssl4eo} & 2022 & \{SAR, Opt\}(10m) & 250K & 264$\times$264 & Geographical-level & $\times$ \\
UBCv2 \cite{huang2023urban} & 2023 & SAR(1m), Opt(0.5m) & 11K & 512$\times$512 & Pixel-level & $\times$ \\
Globe230k \cite{shi2023globe230k} & 2023 & \makecell[l]{SAR(10m), Opt(1m), \\ NDVI(30m), DEM(30m)} & 232K & 512$\times$512 & Geographical-level & $\times$ \\
SkySense \cite{guo2024skysense} & 2024 & \{SAR, MS\}(10m), Opt(0.3m) & 21.5M & \makecell[c]{2048$\times$2048 (High-Res) \\ 64$\times$64 (Low-Res)} & Geographical-level & $\times$ \\
MultiResSAR \cite{zhang2025multi} & 2025 & \{SAR, Opt\}(0.16m-10m) & 11K & 512$\times$512 & Pixel-level & \checkmark \\
MapData \cite{wu2025mapglue} & 2025 & \{Map, Opt\}(1m) & 120K & 512$\times$512 & Pixel-level & $\times$ \\
EarthMiss \cite{zhou2026remote} & 2025 & \{SAR, Opt\}(0.6m), Semantic Masks & 3K & 1024$\times$1024 & Pixel-level & $\times$ \\
3MOS \cite{ye20253mos} & 2025 & \{SAR, Opt\}(3.5m-12.5m) & 113K & 256$\times$256 & Pixel-level & \checkmark \\
MaRS-16M \cite{yang2026mars} & 2025 & \{SAR, Opt\}(0.35m) & 16.8M & 512$\times$512 & Pixel-level & $\times$ \\
\midrule
\textbf{SOMA-1M} & \textbf{2026} & \makecell[l]{\textbf{\{SAR, Opt\}(0.5m, 3m, 10m), Cloud, Geo-coordinates}} & \textbf{1.3M} & \textbf{512$\times$512} & \textbf{Pixel-level} & \textbf{\checkmark} \\
\bottomrule
\end{tabular}
}
\vspace{1ex} 
    \parbox{\textwidth}{ 
        \footnotesize 
        \textbf{Note:} The notation $\{A, B\}(X \text{ m})$ denotes that modalities $A$ and $B$ share a spatial resolution of $X$ m.
    }
\end{table*}

\subsection{Multimodal Remote Sensing Datasets}

Multimodal remote sensing datasets constitute a fundamental data basis for developing general-purpose remote sensing foundation models and supporting downstream applications requiring fine-grained analysis. Although the open-source community has released several SAR-optical datasets in recent years, existing data resources exhibit clear structural polarization, predominantly clustering at two extremes: large-scale low-resolution or small-scale high-resolution. The field currently lacks a unified data framework capable of simultaneously achieving million-scale volume, broad multi-resolution coverage, and accurate pixel-level alignment. To facilitate a structured comparison of existing datasets, we survey representative multimodal remote sensing datasets, as summarized in Table \ref{tab:soma_comparison}, and group them into three categories for discussion.

\subsubsection{Low-to-Mid Resolution Datasets}

This category, represented by SEN12MS \cite{schmitt2019sen12ms}, SEN12MS-CR \cite{ebel2020multisensor}, So2Sat LCZ42 \cite{zhu2020so2sat}, and SSL4EO-S12 \cite{wang2023ssl4eo}, is primarily composed of Sentinel-1 (SAR) and Sentinel-2 (Optical) imagery. Benefiting from the global coverage of Sentinel missions, these datasets typically reach large scales (often exceeding 100,000 image pairs) and are well suited for macro-level semantic tasks, such as land cover classification. However, these datasets exhibit notable limitations in spatial detail and geometric accuracy. First, the 10 m spatial resolution constrains the representation of fine-scale structures such as building footprints, vehicles, and road markings, limiting their applicability to high-precision mapping tasks. Although WHU-OPT-SAR \cite{li2022mcanet} improves the spatial resolution to 5 m, its relatively small data volume is insufficient to support the training of large-capacity models. More importantly, most low-to-mid resolution datasets rely on coarse geo-coordinate alignment and do not explicitly address the geometric distortions introduced by SAR side-looking imaging, particularly in regions with complex terrain. Such distortions often lead to local misalignment, which degrades performance in pixel-sensitive tasks, including image matching and image fusion.

\subsubsection{High-Resolution Datasets}

To support fine-grained perception tasks, datasets such as SpaceNet 6 \cite{shermeyer2020spacenet} (0.5 m), OSdataset (1 m) \cite{xiang2020automatic}, QXS-SAROPT \cite{huang2021qxs}, UBCv2 (0.5 m) \cite{huang2023urban}, MapData (1 m) \cite{wu2025mapglue}, EarthMiss (0.6 m) \cite{zhou2026remote}, and MaRS-16M (0.35 m) \cite{yang2026mars} (0.35 m) provide meter-level or sub-meter aligned imagery. These datasets contain rich texture information and relatively accurate pixel-level registration, making them suitable for tasks such as building extraction and multimodal matching. However, due to the high acquisition costs associated with commercial high-resolution SAR sensors (e.g., Capella, TerraSAR-X), most of these datasets, including UBCv2 and EarthMiss, remain limited in scale and geographic diversity. Such limited spatial coverage and scene diversity increase the risk of model overfitting, thereby constraining generalization to unseen urban environments or geomorphic conditions. Even MaRS-16M, despite its large scale, does not provide explicit multi-resolution support and is primarily designed for single-resolution scenarios.

\subsubsection{Emerging Multi-Res and Foundation Datasets}

In recent years, driven by the growing demand for large-capacity models, the community has begun exploring the construction of larger-scale and more complex hybrid multimodal datasets. Although Globe230k \cite{shi2023globe230k} and SkySense \cite{guo2024skysense} integrate data from multiple sources, they primarily consist of loosely coupled combinations of high-resolution optical imagery and low-resolution SAR imagery, with alignment performed only at the geo-coordinate level. The absence of refined pixel-level registration limits their applicability to cross-modal pixel-level tasks, such as super-resolution or image translation. Regarding multi-resolution alignment, MultiResSAR \cite{zhang2025multi} provides data spanning resolutions from 0.16 m to 10 m; however, with only 11,000 image pairs, its scale is insufficient to support large-scale model pre-training. While 3MOS \cite{ye20253mos} expands the dataset size to approximately 100,000 pairs, its spatial resolution remains within the range of 3.5 m to 12.5 m, lacking sub-meter detail. In addition, its relatively small image size ($256 \times 256$) limits the effective modeling of broader spatial context.

\subsubsection{Summary and Positioning of SOMA-1M}

In summary, existing multimodal remote sensing datasets have achieved notable progress, yet most of them prioritize only one or two aspects of the “scale–resolution–alignment accuracy” triangle. Ultra-large, multi-source datasets such as SkySense emphasize large-scale coverage, but many provide pairing mainly at the geo-coordinate level, making it difficult to guarantee locally consistent pixel-wise geometric correspondence. In contrast, high-quality high-resolution datasets such as MaRS-16M typically offer stronger local registration quality, but they are often tailored to single-resolution settings and lack a unified framework for multi-resolution alignment and cross-scale evaluation. By comparison, the key advantage of SOMA-1M lies in simultaneously achieving million-scale coverage, pixel-level alignment, and explicit multi-resolution support (0.5/3/10 m). This makes SOMA-1M particularly suitable for alignment-sensitive cross-modal pixel-level tasks, such as image matching, fusion, and translation, and enables systematic investigation of cross-scale generalization. Furthermore, SOMA-1M includes an optical degradation subset with simulated clouds spanning diverse morphologies and thicknesses, providing a controllable benchmark for SAR-assisted cloud removal and surface reconstruction. Meanwhile, the dataset preserves accurate geographic coordinates for the center of each sample, forming a structured “SAR–Optical–Geolocation” triplet that supports research on location encoding and multimodal contrastive pre-training. Overall, SOMA-1M integrates million-scale coverage with explicit multi-resolution support and pixel-level alignment, and further provides task-specific subsets and geolocation metadata, making it well suited for multimodal remote sensing research and downstream evaluation.

\subsection{Multi-Modal Applications}

The all-weather geometric structural information provided by SAR imagery and the rich spectral and textural features offered by optical imagery are complementary in nature. This complementarity forms the basis for a range of downstream multimodal remote sensing tasks, including image matching, image fusion, SAR-assisted cloud removal, and image translation.

\subsubsection{Image Matching}

Image matching is a fundamental step for multimodal data fusion, registration, and 3D reconstruction \cite{zhang2025soft, zhang2023histogram, zheng2025msg, zhang2025glift, lowe2004distinctive}. Due to significant non-linear radiometric differences (NRD) and complex geometric distortions between SAR and optical imagery, achieving high-precision multimodal matching remains a challenging problem in the remote sensing field. Existing research generally falls into two categories, namely traditional methods based on hand-crafted features and data-driven deep learning methods.

Traditional methods. To overcome the failure of gradient-based features such as SIFT \cite{lowe2004distinctive} on multimodal imagery, early studies focused on designing structural descriptors that are robust to radiometric variations. Representative works include HOWP \cite{zhang2023histogram}, which addresses gradient reversal through weighted phase histograms; SOFT \cite{zhang2025soft}, which introduces second-order tensor representations to handle non-rigid deformations; MSG \cite{zheng2025msg}, which employs side-window Gaussian filtering to extract stable edge features; and GLIFT \cite{zhang2025glift}, which combines a global–local search strategy with 3D descriptors to cope with large rotation and scale variations. However, these methods primarily rely on shallow local geometric cues and often fail to establish reliable semantic correspondences in scenes with repetitive structures (e.g., dense building areas) or large viewpoint changes. In addition, their iterative optimization procedures incur high computational cost, limiting scalability to large-scale datasets.

Deep Sparse Matching. With the development of deep learning, the detector-descriptor-matcher pipeline has gradually dominated this field. In general computer vision, attention-based matchers such as SuperGlue \cite{sarlin2020superglue} and LightGlue \cite{lindenberger2023lightglue}, as well as recent diffusion-based DiffGlue \cite{zhang2024diffglue} and state-space model-based MambaGlue \cite{11128473}, have all achieved progress on large-scale natural image datasets like MegaDepth \cite{li2018megadepth}. Even for methods optimized for multimodal tasks, such as MINIMA \cite{ren2025minima}, their training data remains based on multimodal variants constructed from MegaDepth (e.g., RGB-Depth or Day-Night), which essentially fall within the category of natural scenes. Due to the fundamental differences in imaging mechanisms between natural images and SAR imagery, directly transferring these pre-trained models to SAR-optical tasks often faces severe Domain Shift problems. To address this, dedicated methods in remote sensing, such as ReDFeat \cite{deng2022redfeat} and MapGlue \cite{wu2025mapglue}, utilize metric learning or semantic guidance strategies to extract modality-invariant features. However, constrained by data resources, these methods are typically trained only on small-scale (e.g., OSdataset) or medium-scale (e.g., MapData) remote sensing datasets. The lack of support from large-scale diverse samples makes it difficult for these models to generalize to unseen complex terrains.

From Semi-Dense to Dense Matching: To address the scarcity of sparse feature points in low-texture regions, Transformer-based Detector-free methods have emerged. LoFTR \cite{sun2021loftr} achieved semi-dense matching by establishing pixel-level long-range dependencies, while its subsequent variants, ELoFTR \cite{wang2024efficient} and XoFTR \cite{tuzcuouglu2024xoftr}, further optimized local detail perception and efficiency. Regarding dense matching, RoMa \cite{edstedt2024roma} achieved pixel-level full-image matching via robust Markov regularization. However, all the aforementioned semi-dense and dense methods are currently trained primarily on the MegaDepth dataset. While they exhibit excellent performance in natural scenes, transferring their capabilities to the SAR-optical domain strictly requires re-training or fine-tuning on large-scale remote sensing data with high geometric alignment accuracy. Existing multimodal remote sensing datasets (e.g., SEN12MS) typically possess only coarse registration at the geo-coordinate level. The presence of pixel-level errors prevents models like RoMa from learning precise texture mapping relationships. This directly results in the potential of such powerful dense matching models remaining fully unexplored in SAR-optical tasks.

\subsubsection{Image Fusion}

Image fusion aims to integrate complementary information from multi-source sensors to generate synthetic imagery with improved scene representation and interpretability. In the computer vision community, deep learning-based fusion frameworks have become the dominant paradigm, with most developments concentrating on infrared–optical fusion in natural scenes. These methods typically follow a data-driven design philosophy and have demonstrated strong performance under well-controlled conditions. Early research focused on constructing end-to-end feature mapping architectures; for instance, RFN-Nest \cite{li2021rfn} utilized a residual-based nested connection network to achieve automated multi-scale feature extraction and reconstruction. Subsequently, to address the "black-box" nature of deep networks and its inherent lack of interpretability, LRRNet \cite{li2023lrrnet} innovatively unfolded Low-Rank Representation optimization algorithms into deep networks, while MUFusion \cite{cheng2023mufusion} introduced memory units to establish a self-evolving training paradigm, enhancing the feature representation capabilities of unsupervised learning. More recent studies have further branched into algorithmic efficiency and adaptation to complex environments: for nighttime or harsh conditions, LENFusion \cite{chen2024lenfusion} innovatively coupled the fusion task with low-light enhancement; to meet real-time inference demands, MMDRFuse \cite{deng2024mmdrfuse} and LUT-Fuse \cite{yi2025lut} achieved lightweight, ultra-fast model inference using knowledge distillation and learnable look-up tables, respectively. Additionally, to overcome the loss of high-frequency details during the fusion process, RPFNet \cite{guan2025residual} introduced a residual prior and frequency-aware mechanism, further enhancing the model's ability to capture minute textures through cross-domain frequency interaction.

In the field of remote sensing, the objective of image fusion is to synergistically leverage the unique geometric structural information of SAR imagery and the rich spectral textures of optical imagery. However, unlike the flourishing state of the infrared-optical domain, there remains a notable lack of deep learning fusion models specifically designed for the multimodal mechanisms of SAR and optical data. Consequently, research in this area still largely relies on traditional methods with sound mathematical interpretability. Representative works include VSFF \cite{ye2024optical}, which proposed a visual saliency framework based on complementary feature decomposition using total variation constraints to reconstruct salient features, and BASHVS \cite{gong2025bashvs}, which further introduced a human visual system attention mechanism and utilized synchronous anisotropic diffusion for feature aggregation. Although these traditional methods are theoretically robust, they rely on hand-crafted rules (e.g., weighted averaging) that struggle to adaptively handle SAR speckle noise and complex spectral differences.

Limitations of Cross-Domain Transfer. Due to the lack of dedicated models, researchers often attempt to directly transfer the aforementioned infrared-optical deep models to SAR-optical tasks, which faces two severe challenges: (1) Significant Domain Shift: The coherent speckle noise and side-looking imaging characteristics of SAR are fundamentally different from the thermal radiation properties of infrared images. Directly reusing infrared-optical pre-trained weights often leads to misinterpretation of SAR structures or loss of texture. (2) Extreme Alignment Sensitivity: Existing infrared-optical models typically assume that input images are perfectly registered. However, in practical SAR-optical imagery, minute misalignments caused by terrain variations are extremely common. In the absence of pixel-level aligned training data, the fusion results generated by these models are highly prone to ghosting artifacts or edge blurring, which severely constrains the application of deep fusion technology in high-precision mapping.

\subsubsection{SAR-Assisted Cloud Removal}

SAR-assisted cloud removal leverages the penetrative capability of SAR imagery to recover optical information obscured by cloud cover, which can be formulated as a structure-constrained texture reconstruction task. In contrast to single-modality cloud removal methods, which typically rely on local neighborhood context for image inpainting, SAR data provides explicit geometric cues of underlying ground structures, offering reliable structural guidance for cloud-obscured regions. Such structural information plays an important role in preserving geometric consistency in the reconstructed optical images. Existing methods have gradually evolved from discriminative feature fitting to generative reconstruction frameworks. Early CNN-based approaches, such as Dsen-CR \cite{meraner2020cloud}, constructed multimodal cloud removal pipelines by cascading SAR and optical features within deep residual networks. GLF-CR \cite{xu2022glf} further introduced a Global–Local Fusion strategy to alleviate feature incompatibility caused by large modal differences. To mitigate spatial misalignment commonly observed in multi-source remote sensing data, Align-CR \cite{xu2023multimodal} incorporated feature alignment modules to correct geometric deviations in latent space, reducing structural artifacts induced by misregistration. ACA-CRNet \cite{huang2024attentive} subsequently introduced an attentive contextual attention mechanism to suppress SAR speckle noise and discard irrelevant auxiliary features. To address the limited receptive field of convolutional architectures, HPN-CR \cite{gu2025hpn} integrated Transformer-based global modeling with heterogeneous parallel networks, enabling improved reconstruction of large cloud-covered regions with reduced color distortion and texture inconsistency. More recently, diffusion-based generative models have been explored for SAR-assisted cloud removal. EMRDM \cite{liu2025effective} adopted a mean reversion-driven direct diffusion framework, demonstrating improved realism and fine-detail preservation in reconstructed optical textures. Despite continuous architectural advancements, experimental evaluation in this field remains constrained by data resolution. Most existing methods are trained and validated on the SEN12MS-CR dataset (Sentinel-1/2, 10 m GSD), where the coarse spatial resolution limits the representation of high-frequency structures such as vehicles, road markings, and densely distributed buildings. At this resolution, SAR guidance mainly captures large-scale land-cover boundaries rather than fine-grained structural details, restricting the assessment of SAR-assisted cloud removal in complex urban scenarios. The absence of multi-resolution, pixel-level aligned SAR–optical benchmarks has therefore hindered the exploration of medium- and high-resolution SAR for cloud removal tasks.

\subsubsection{SAR-to-Optical Translation}

SAR-to-optical translation aims to map SAR imagery to the optical domain, with the objective of preserving geometric structures while generating visually plausible optical-like textures. This task involves a complex cross-modal mapping process, as models must compensate for geometric distortions caused by side-looking SAR imaging and suppress coherent speckle noise, while inferring optical appearance from limited microwave scattering responses. According to the availability of paired training data, existing approaches can be broadly categorized into unsupervised and supervised methods.

Unsupervised methods aim to achieve cross-modal translation without paired SAR–optical samples. Early studies primarily relied on GAN-based frameworks. CycleGAN \cite{zhu2017unpaired} introduced cycle-consistency loss to constrain bidirectional mappings between source and target domains. CUT \cite{park2020contrastive} employed contrastive learning to enable efficient unidirectional translation by maximizing patch-level mutual information. To better address SAR-specific noise characteristics and cloud-contaminated regions, MT\_GAN \cite{wang2025mt_gan} modified the adversarial architecture for SAR-to-optical translation and cloud removal. More recently, diffusion-based models have been introduced into this task. CycleDiffusion \cite{wu2023latent} proposed a framework based on DPM-Encoders, enabling unpaired translation by aligning the latent spaces of independently trained diffusion models. UNSB \cite{kim2024unpaired} formulated SAR-to-optical translation as a Schrödinger bridge problem and solved the corresponding stochastic differential equations through adversarial training. CM-Diff \cite{hu2025cm} further extended this line of work by modeling bidirectional cross-modal translation within a unified diffusion framework, allowing simultaneous approximation of SAR and optical data distributions. Despite their reduced reliance on paired data, unpaired methods often lack explicit pixel-level geometric constraints, which can result in noticeable geometric distortions when translating complex man-made structures such as dense urban areas.

Supervised methods, in contrast, enforce structural consistency through pixel-level alignment between SAR and optical image pairs. pix2pix \cite{isola2017image} proposed a conditional GAN framework with L1 and adversarial losses to constrain pixel-wise correspondence, while pix2pixHD \cite{wang2018high} further extended it to high-resolution image generation. BicycleGAN \cite{zhu2017toward} introduced bijective consistency constraints to alleviate mode collapse and enable diverse outputs from a single input. With the development of diffusion models, BBDM \cite{li2023bbdm} modeled the translation process as a Brownian Bridge, directly connecting source and target domains without relying on standard Gaussian noise schedules. To improve inference efficiency, ACD\_S2ODPM \cite{bai2024accelerating} combined adversarial consistency distillation with diffusion-based generation, achieving faster inference while maintaining translation quality. Although supervised approaches generally achieve superior geometric fidelity, their effectiveness strongly depends on the quality of paired training data. Existing public datasets often suffer from limited spatial resolution or imperfect registration, which hinders accurate learning of the mapping from SAR scattering patterns to optical textures. Therefore, high-resolution datasets with strict pixel-level alignment, such as SOMA-1M, are essential for training reliable supervised SAR-to-optical translation models.

In summary, although multimodal remote sensing model architectures continue to emerge, their performance upper bound remains constrained by the scarcity of high-quality, multi-resolution aligned data. The introduction of SOMA-1M aims to break this core bottleneck at the data level, and we will detail the construction process of SOMA-1M in the next chapter.

\begin{figure*}[!tbp] 
    \centering
    \includegraphics[width=1.00\textwidth]{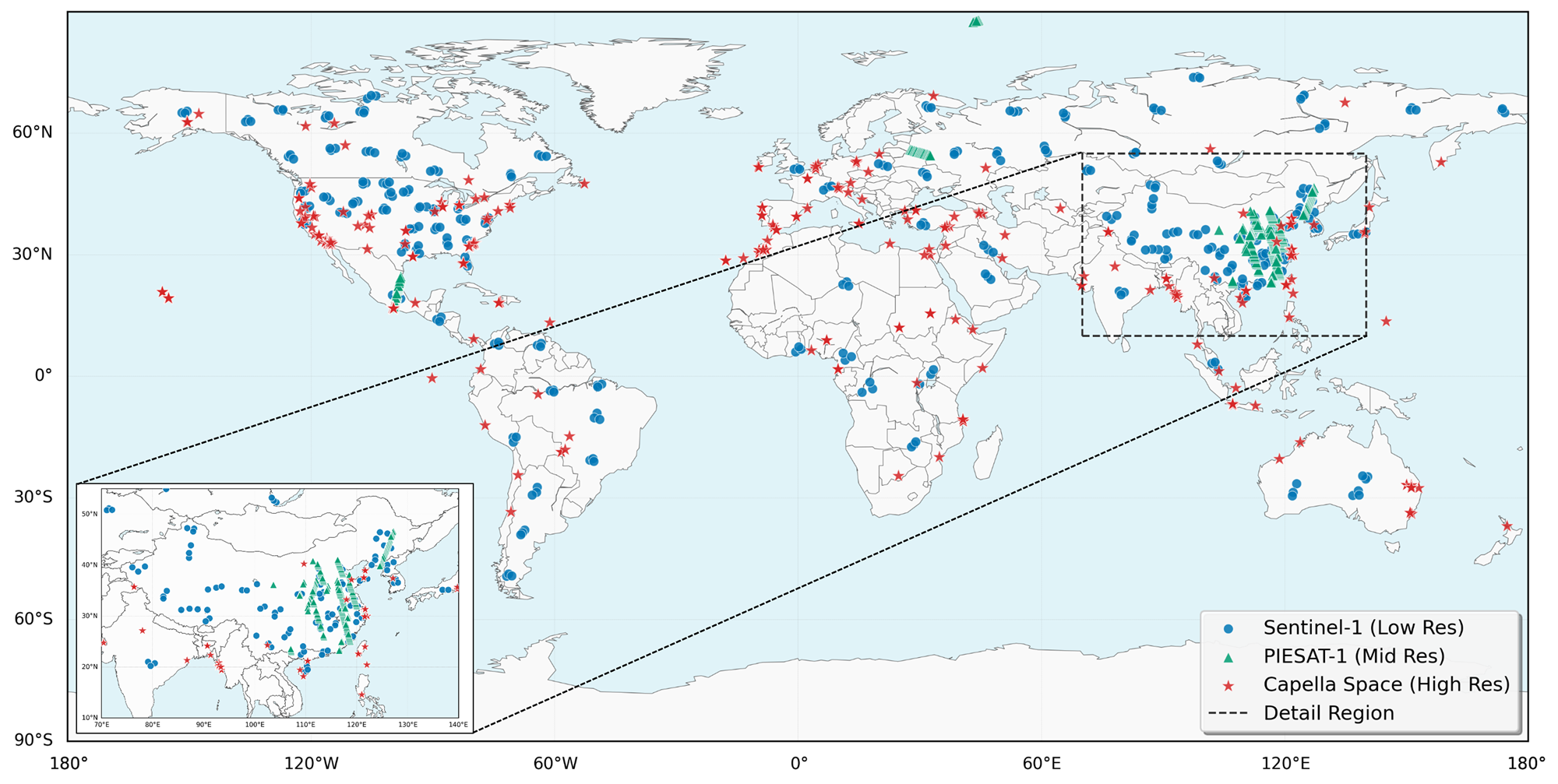}
    \caption{Global geographic distribution of SOMA-1M sampling points.}
    \label{fig:Sampling Points}
\end{figure*}

\begin{figure*}[!t] 
    \centering
    \includegraphics[width=1.00\textwidth]{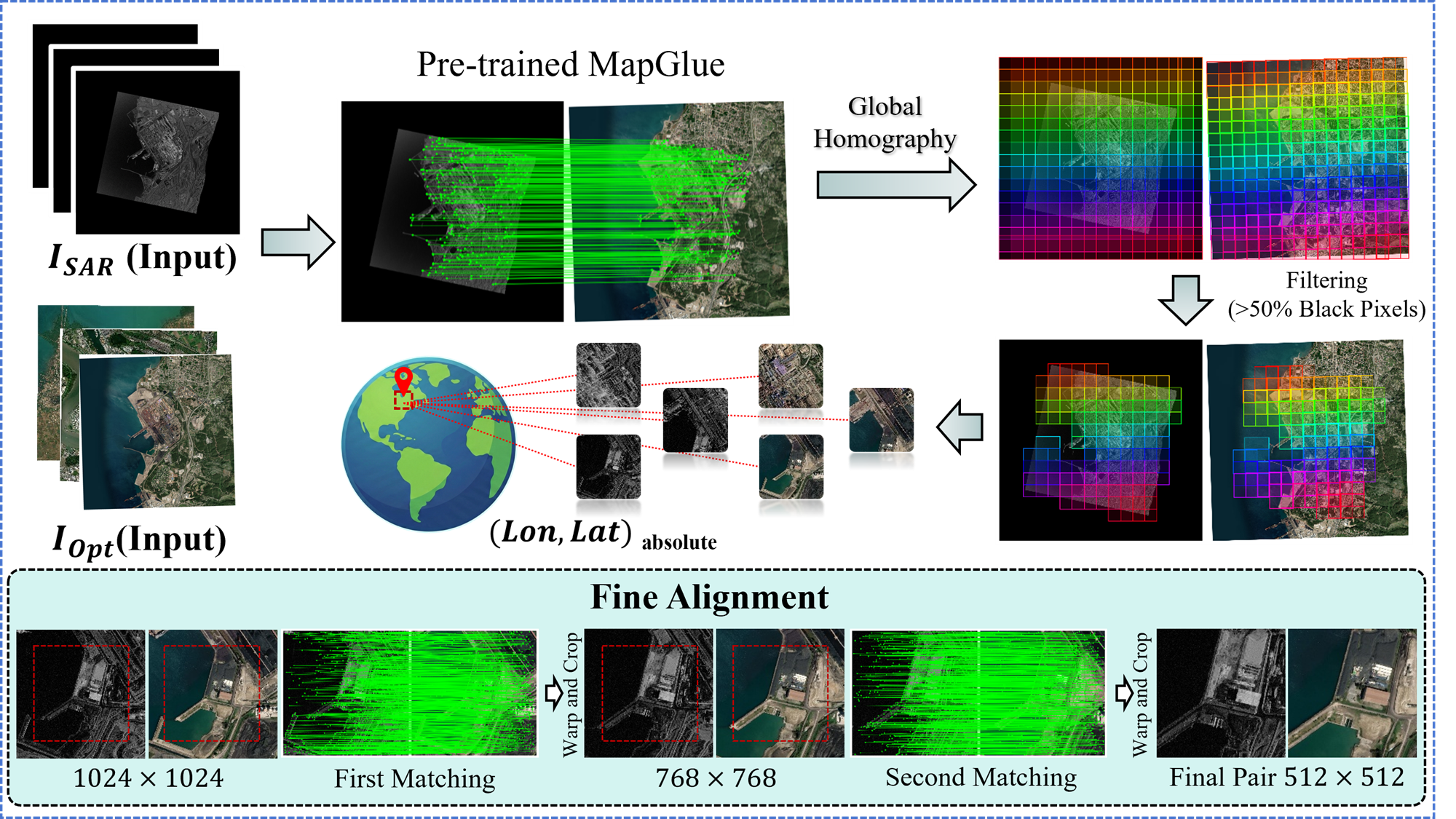}
    \caption{Flowchart of the automated data annotation pipeline.}
    \label{fig:annotation pipeline}
\end{figure*}

\section{SOMA-1M Dataset Construction}

The construction of the SOMA-1M dataset was a systematic engineering project spanning 10 months (December 2024 to October 2025). The project involved the coordinated efforts of more than 20 remote sensing professionals, who collaboratively conducted data surveying, downloading, processing, and quality cleaning, thereby ensuring the high quality and reliability of the dataset.

\subsection{Data Collection}

To cover a wide range of spatial resolutions and sensor characteristics, we constructed a global SAR–optical imagery library, as shown in Fig. \ref{fig:Sampling Points}. Optical imagery was primarily sourced from Google Earth historical imagery, while SAR imagery was grouped into three resolution levels according to sensor type and spatial resolution:

\begin{itemize}
    \item Low resolution: Sentinel-1 SAR imagery at 10 m resolution paired with 8 m resolution Google Earth optical imagery, comprising 343 original scenes and producing 357,563 valid patches.
    \item Mid-resolution: PIESAT-1 SAR imagery at 3 m resolution paired with 4 m resolution Google Earth optical imagery, comprising 628 original scenes and producing 834,265 valid patches.
    \item High-resolution: Capella Space SAR imagery at 0.5 m resolution paired with 1 m resolution Google Earth optical imagery, comprising 495 original scenes and producing 109,126 valid patches.
\end{itemize}

Data collection covered 1,466 geographic locations worldwide, resulting in a total of 1,300,954 SAR–optical image pairs across representative land-cover types, including urban areas, rural regions, mountainous terrain, river basins, and deserts. The spatial resolution of the original scenes ranges from 0.5 m to 10 m, with individual image sizes ranging from 8,000 to 35,000 pixels. As illustrated by the sampling distribution in Fig. \ref{fig:Sampling Points}, low- and high-resolution samples are globally distributed, whereas mid-resolution samples are primarily concentrated in East Asia.

\begin{figure*}[!t] 
    \centering
    \includegraphics[width=1.00\textwidth]{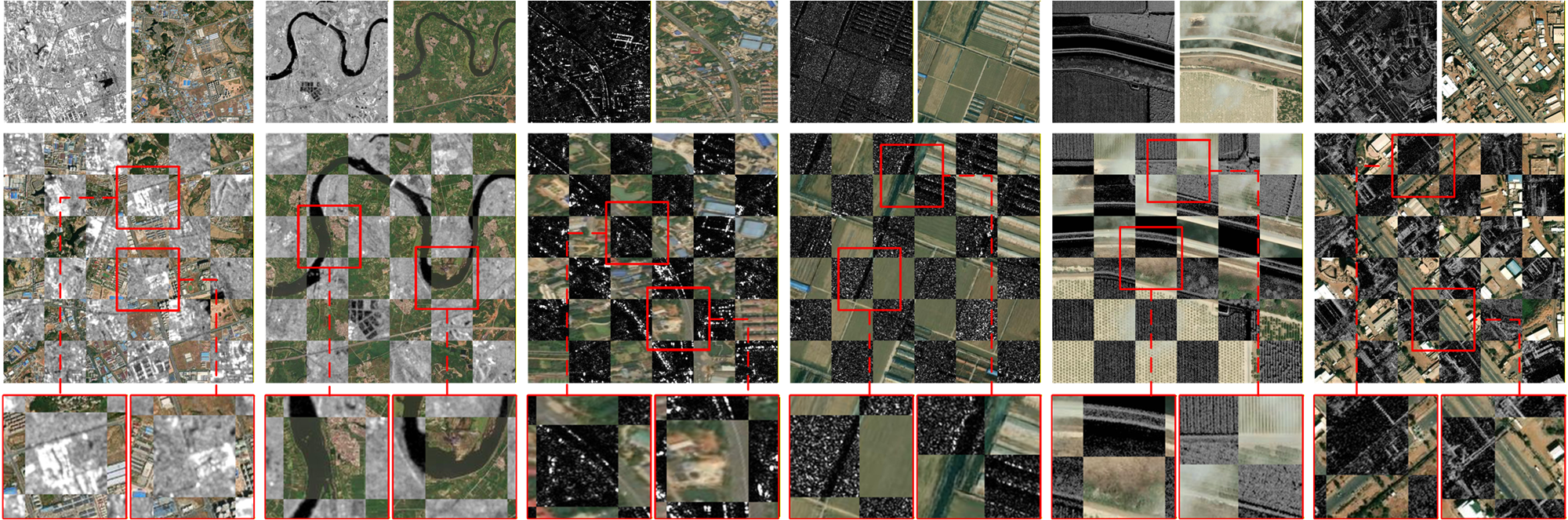}
    \caption{Visualization of alignment results.}
    \label{fig:Visualization of alignment results}
\end{figure*}

\begin{figure*}[!t] 
    \centering
    \includegraphics[width=1.00\textwidth]{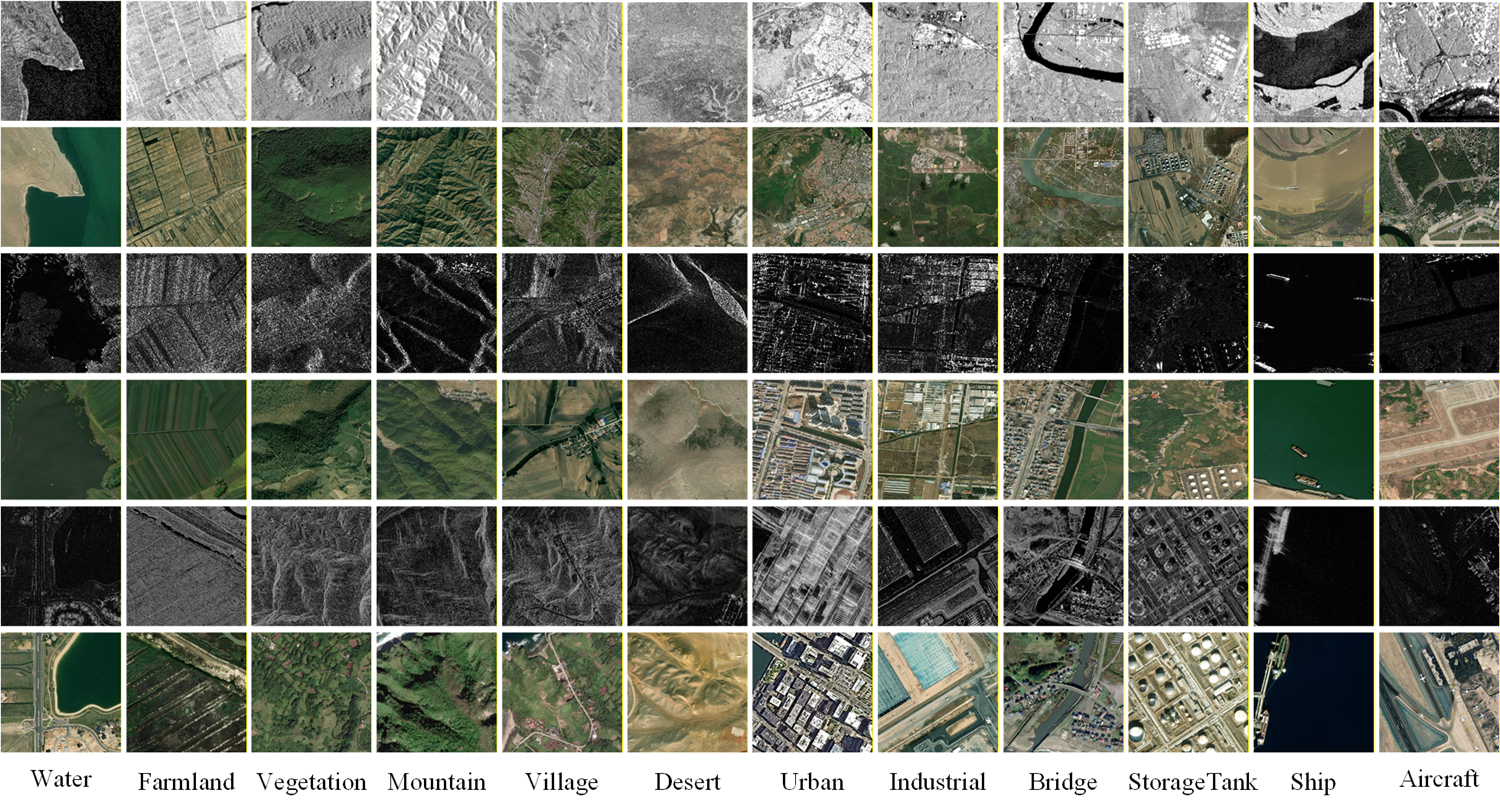}
    \caption{Visualization examples of 12 typical land-cover categories in the SOMA-1M dataset. Each group presents a pair of SAR and optical images with strict pixel-level alignment.}
    \label{fig:typical categories}
\end{figure*}

\subsection{Automated Data Annotation}

To address the challenges of cross-modal registration between heterogeneous imagery (SAR and optical), this study designed a fully automated, two-stage alignment pipeline (as illustrated in Fig. \ref{fig:annotation pipeline}). The core of this framework lies in leveraging the cross-modal matching capability of the MapGlue model. MapGlue is a matching model pre-trained on a dataset comprising 120,000 pairs of electronic maps and optical imagery; although trained on a different modality pair, experiments demonstrate that it can be effectively applied to SAR–optical image matching. The two-stage annotation process is as follows:

Coarse Alignment. We first downsample the original large-scale images to 1024 $\times$ 1024 pixels and apply MapGlue to extract 4,096 keypoints for global matching. The RANSAC threshold is set to 10 pixels to estimate the global transformation matrix. The original optical image is then warped into the coordinate system of the original SAR image to complete the coarse alignment. Subsequently, the aligned large-scale images are sliced into patches with a size of 1,024 and a stride of 612. To handle potential black regions in the image tiles, a pixel count is performed for each patch; if the black region exceeds 50\%, the image block is discarded.

Fine Alignment. To achieve pixel-level accuracy, we perform a cascaded local refinement procedure on the sliced patches. During the first refinement stage, matching is performed again on the 1024 $\times$ 1024 image blocks, with the RANSAC threshold tightened to 1.5 pixels to estimate the local homography matrix $\boldsymbol{H}$ and execute a secondary warp operation. To eliminate edge distortions and black regions resulting from the transformation, we crop a 768 $\times$ 768 image block from the central area. In the second refinement stage, the same matching and warping procedure is repeated on the 768 $\times$ 768 image blocks, and a final 512 $\times$ 512 image block is cropped from the center as the definitive sample, with a maximum overlap of 100 pixels between samples. Through this multi-level refined matching strategy, we obtained SAR–optical image pairs with pixel-level alignment accuracy. As shown in Fig. \ref{fig:Visualization of alignment results}, pixel-level alignment is achieved across diverse scenes, including urban streets, farmlands, rivers, and bridges.

We conducted a manual quality inspection on 100,000 randomly selected sample pairs from the produced dataset, yielding a qualification rate exceeding 99.8\%. Even in regions with complex terrain and pronounced non-linear geometric distortions, the image pairs exhibit high visual consistency, as shown in Fig. \ref{fig:typical categories}. Furthermore, we preserved the absolute geo-coordinates of the center point for each image block during the production process. This enables SOMA-1M to support geolocation-based contrastive learning (e.g., SatCLIP \cite{klemmer2025satclip}) to train location encoders for geo-coordinates, thereby supporting downstream tasks such as temperature prediction, population density estimation, and species classification.

\begin{figure*}[!t] 
    \centering
    \includegraphics[width=1.00\textwidth]{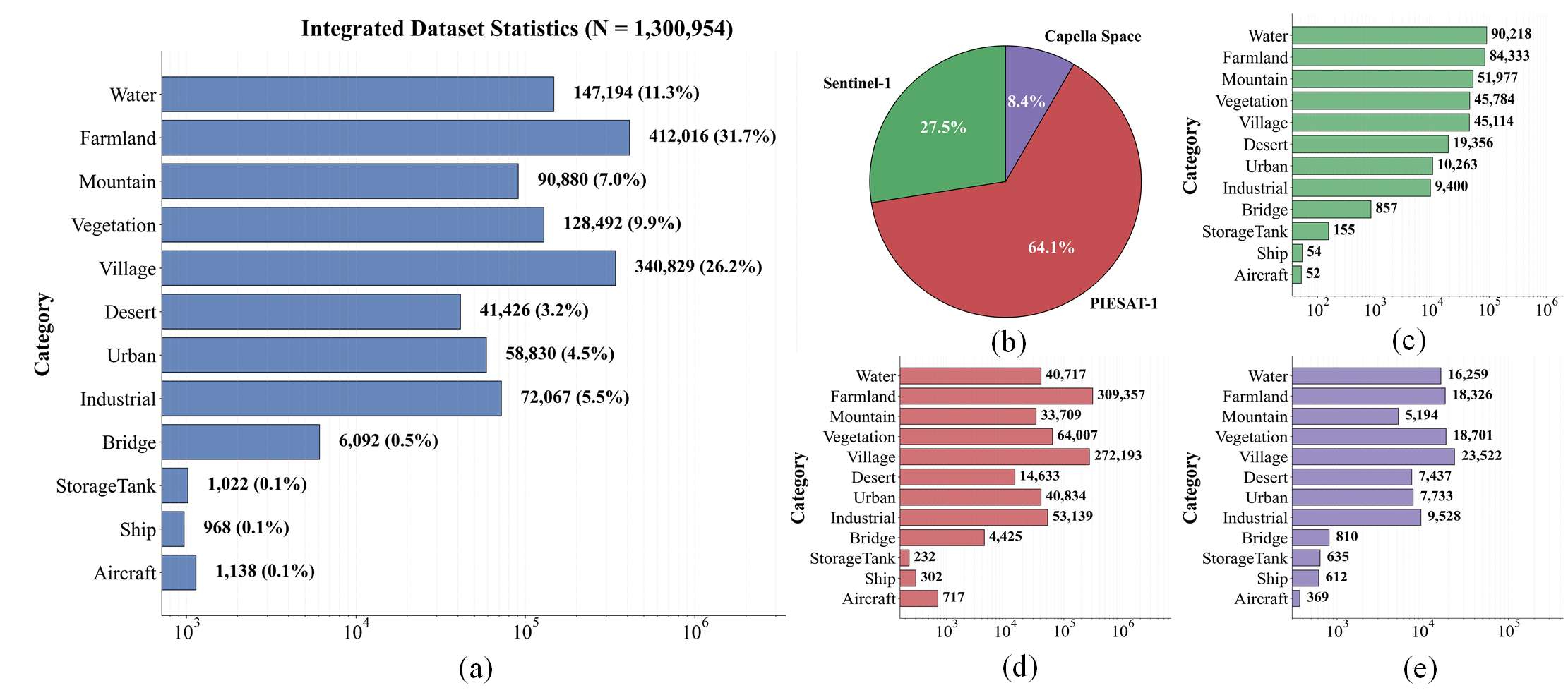}
    \caption{Category statistics and sensor distribution ratios for the SOMA-1M dataset. (a) Distribution of categories for all samples; (b) proportions of the three resolution subsets; (c)-(e) detailed category statistics for the Sentinel-1, PIESAT-1, and Capella Space subsets, respectively.}
    \label{fig:Category statistics}
\end{figure*}

\begin{figure*}[!t] 
    \centering
    \includegraphics[width=1.00\textwidth]{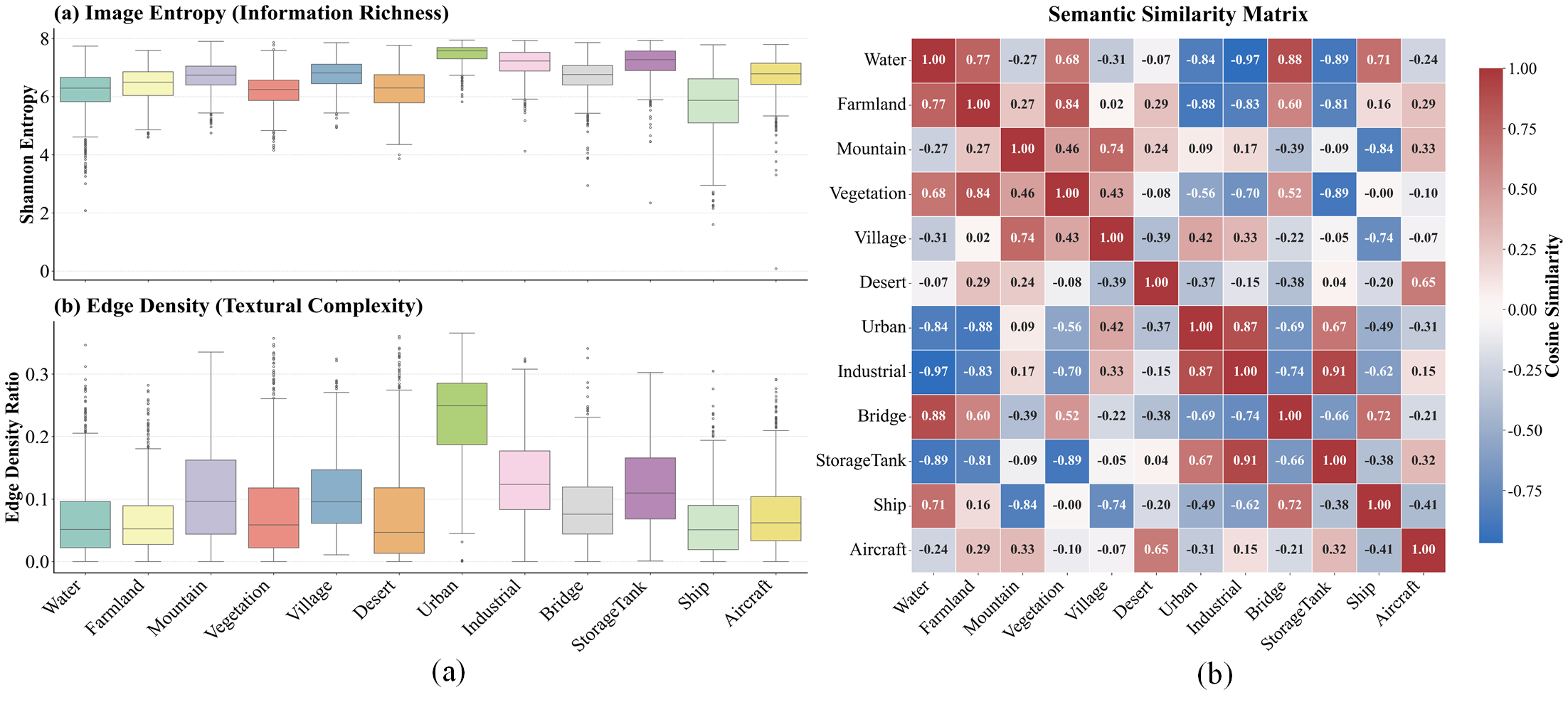}
    \caption{Multi-dimensional characteristic analysis of the SOMA-1M dataset. (a) Physical complexity of scenes; (b) semantic similarity of categories.}
    \label{fig:Multi-dimensional}
\end{figure*}

\subsection{Data Analysis}

SOMA-1M contains a total of 1,300,954 strictly aligned multimodal image patches, covering 12 typical semantic labels across three hierarchical levels, namely high-value targets, man-made settlements, and natural backgrounds (Fig. \ref{fig:typical categories}).

\subsubsection{Sample Scale and Multi-scale Distribution}

Based on the overall statistical results shown in Fig. \ref{fig:Category statistics}a, the dataset exhibits a long-tail distribution characteristic consistent with real-world remote sensing applications, where farmland ($31.7\%$) and villages ($26.2\%$) constitute the bulk of the dataset, reflecting the most common land-cover scenarios globally. In contrast, while the proportion of high-value small target samples such as oil tanks, ships, and aircraft is relatively small (approximately $0.1\%$), their absolute quantities remain at the scale of thousands, thereby supporting feature mining for rare objects by deep models. By integrating data from three types of sensors—Sentinel-1, PIESAT-1, and Capella Space (Fig. \ref{fig:Category statistics}b)—the dataset achieves a multi-scale balance ranging from global wide-area coverage to local fine-grained sampling. Specifically, the Sentinel-1 subset shows a significant proportion in water body classification (Fig. \ref{fig:Category statistics}c), while PIESAT-1 contributes the highest data density for farmland and village categories (Fig. \ref{fig:Category statistics}d). Capella Space focuses on providing high-precision ground object structures in urban and industrial areas (Fig. \ref{fig:Category statistics}e). This complementarity across resolution levels supports the evaluation of cross-resolution feature learning and scene generalization.

\subsubsection{VLM-based Automated Classification}

To ensure accuracy and logical consistency across large-scale sample classification, this study utilizes the vision-language model Qwen3-VL-8B \cite{Qwen3-VL} to execute an automated annotation process. This process follows a rule-based semantic hierarchy inspired by common remote sensing interpretation practices, where high-value targets (e.g., aircraft, ships, bridges, oil tanks) take precedence over human settlements (urban, village, industrial), which in turn take precedence over natural backgrounds (mountains, vegetation, farmland, water bodies, deserts). To resolve the challenge of mixed land cover, the system employs refined discrimination rules; for example, when rural buildings and cropland coexist, the algorithm prioritizes the settlement features to classify the scene as village rather than farmland, thereby maintaining semantic consistency in complex geographical contexts.

\subsubsection{Multi-dimensional characteristic analysis}

To analyze the intrinsic characteristics of the SOMA-1M dataset, this study systematically evaluates its scene complexity and semantic correlation through physical index quantification and semantic feature decoupling, as shown in Fig. \ref{fig:Multi-dimensional}. Assessments utilizing information entropy and edge density reveal that man-made features, such as urban and industrial areas, demonstrate higher median entropy values (mean > 7.2) and edge density due to their complex spectral responses and dense high-frequency geometric structures. In contrast, homogeneous land covers such as water bodies and deserts exhibit comparatively low structural complexity. This broad spectral and multi-textured gradient coverage, ranging from weak-textured scenes to high-frequency complex environments, ensures that SOMA-1M provides diverse and challenging samples for multimodal alignment algorithms. Furthermore, the semantic similarity matrix based on image statistical features shows significant feature coupling between man-made objects such as urban areas, industrial zones, and oil tanks (all similarities > 0.87), while natural categories also maintain strong affinities. Notably, a strong negative correlation (-0.96) is observed between water bodies and industrial areas. This distinct physical disparity indicates that the dataset spans a wide separability range within the feature space. In summary, multi-dimensional statistical analysis demonstrates that SOMA-1M constitutes a large-scale multimodal remote sensing dataset with clear semantic organization and substantial physical diversity.

\section{Applications and Benchmarks}

This chapter aims to evaluate the role of the SOMA-1M dataset across multiple hierarchical visual tasks, including geometric alignment, pixel-level processing, semantic understanding, and image generation, through a series of representative downstream experiments. Considering the massive scale of the overall imagery, the full dataset is not used in the experiments; instead, a training subset containing 100,000 representative image pairs, designated as SOMA-0.1M, was selected for model training or fine-tuning. In this subset, image pairs from the Sentinel-1, PIESAT-1, and Capella Space sensors each account for one-third of the total. Results obtained on this subset, which represents less than 10\% of the complete dataset, are used to analyze the impact of pixel-level aligned data on cross-modal representation learning. Furthermore, for low-, middle-, and high-resolution scenarios, 1,000 image pairs were independently sampled for each resolution to construct a benchmark test set totaling 3,000 pairs, referred to as SOMA-Test. This design ensures a balanced and objective evaluation across different spatial resolutions and sensor characteristics. All experiments were conducted on an NVIDIA A100 (80GB) computing platform, providing a consistent experimental environment for all evaluated methods.

\subsection{Image Matching}

This section examines the contribution of the SOMA-1M dataset to geometric feature learning. Specifically, we compare the matching performance of models fine-tuned on a subset of SOMA-1M (denoted as SOMA-0.1M) with that of models trained on natural images or other existing remote sensing datasets. To evaluate the generalization performance across different data distributions, we utilize the following three multi-modal remote sensing imagery test sets:

\textit{SOMA-Test}: The internal test set constructed from the dataset proposed in this paper.

\textit{OSdataset} \cite{xiang2020automatic}: An external dataset containing 662 pairs of Gaofen-3 SAR and optical imagery, including both validation and test samples, used to assess cross-dataset generalization to different SAR sensors.

\textit{SRIF} \cite{li2023multimodal}: A diverse multi-modal dataset comprising 1,200 image pairs across six modalities, including multi-temporal optical, infrared-optical, depth-optical, map-optical, SAR-optical, and day–night imagery. In this dataset, infrared-optical and depth-optical pairs are derived from natural images, while the remaining modalities correspond to remote sensing imagery. Each modality contains 200 pairs, and the full dataset is used to evaluate the zero-shot robustness of the models in other multimodal settings.

\begin{table*}[t]
\centering
\caption{Quantitative evaluation results of different image matching algorithms on SOMA-Test, OSdataset, and SRIF datasets.}
\label{tab:matching_quantitative_eval}

\footnotesize 
\setlength{\tabcolsep}{6pt} 

\begin{tabular}{c|c|c|c|c|c|c|c|c|c|c}
\toprule
\multirow{2.5}{*}{Category} & \multirow{2.5}{*}{Method} & \multicolumn{3}{c|}{SOMA-Test} & \multicolumn{3}{c|}{OSdataset} & \multicolumn{3}{c}{SRIF} \\
\cmidrule(lr){3-5} \cmidrule(lr){6-8} \cmidrule(lr){9-11}
 & & @5px & @10px & @20px & @5px & @10px & @20px & @5px & @10px & @20px \\
\midrule

\multirow{15}{*}{Sparse} 
 & MSG & 5.30 & 9.13 & 12.10 & 4.98 & 14.35 & 20.69 & 10.50 & 16.92 & 21.17 \\
 & GLIFT & 1.42 & 5.62 & 14.58 & 0.15 & 4.83 & 14.80 & 2.25 & 8.17 & 20.10 \\
 & ReDFeat & 0.23 & 0.90 & 1.90 & 0.15 & 0.76 & 1.51 & 0.75 & 2.00 & 2.58 \\
 & $\text{SOMA}_{\text{ReDFeat}}$ & \bv{3.23} & \bv{6.23} & \bv{9.50} & \bv{2.57} & \bv{5.89} & \bv{10.73} & \bv{2.58} & \bv{6.42} & \bv{10.17} \\
 & LightGlue & 0.50 & 1.38 & 2.57 & 0.29 & 1.06 & 2.63 & 3.18 & 4.67 & 7.05 \\
 & $\text{MINIMA}_{\text{LightGlue}}$ & 1.61 & 4.20 & 7.53 & 1.39 & 4.92 & 10.56 & 3.44 & 8.86 & 13.97 \\
 & $\text{SOMA}_{\text{LightGlue}}$ & \bv{6.40} & \bv{18.29} & \bv{33.61} & \bv{3.88} & \bv{18.28} & \bv{40.55} & \bv{5.13} & \bv{16.51} & \bv{29.01} \\
 & MambaGlue & 0.05 & 0.12 & 0.19 & 0.00 & 0.00 & 0.00 & 0.82 & 2.04 & 3.11 \\
 & $\text{SOMA}_{\text{MambaGlue}}$ & \bv{4.65} & \bv{10.41} & \bv{16.77} & \bv{2.91} & \bv{11.23} & \bv{22.45} & \bv{4.87} & \bv{12.38} & \bv{20.25} \\
 & DiffGlue & 0.42 & 1.06 & 1.99 & 0.15 & 0.42 & 1.42 & 1.97 & 4.68 & 7.38 \\
 & $\text{SOMA}_{\text{DiffGlue}}$ & \bv{3.88} & \bv{10.20} & \bv{18.61} & \bv{1.75} & \bv{9.64} & \bv{23.73} & \bv{3.80} & \bv{12.06} & \bv{21.86} \\
 & LightGlueStick & 0.42 & 1.22 & 2.40 & 0.20 & 0.97 & 2.96 & 1.93 & 4.78 & 8.14 \\
 & $\text{SOMA}_{\text{LightGlueStick}}$ & \bv{5.02} & \bv{14.14} & \bv{26.98} & \bv{3.58} & \bv{15.57} & \bv{34.97} & \bv{5.57} & \bv{18.01} & \bv{32.88} \\
 & MapGlue & 11.25 & 24.28 & 38.89 & 8.36 & 27.56 & 49.11 & 13.35 & 30.70 & 45.68 \\
 & $\text{SOMA}_{\text{MapGlue}}$ & \rb{18.63} & \rb{37.25} & \rb{55.42} & \rb{11.27} & \rb{34.70} & \rb{58.23} & \rb{17.23} & \rb{35.42} & \rb{49.72} \\
\midrule

\multirow{6}{*}{\makecell{Semi-\\dense}}
 & XoFTR & 1.24 & 3.11 & 5.02 & 0.87 & 3.34 & 6.09 & 1.99 & 6.13 & 11.55 \\
 & $\text{MINIMA}_{\text{XoFTR}}$ & 0.55 & 1.54 & 2.81 & 0.47 & 1.73 & 3.57 & 1.93 & 6.31 & 11.71 \\
 & $\text{SOMA}_{\text{XoFTR}}$ & \bv{5.84} & \bv{15.26} & \bv{27.08} & \bv{4.19} & \bv{16.78} & \bv{33.52} & 0.76 & 4.00 & 9.19 \\
 & ELoFTR & 0.29 & 0.77 & 1.61 & 0.00 & 0.38 & 1.45 & 0.50 & 2.00 & 4.92 \\
 & $\text{MINIMA}_{\text{ELoFTR}}$ & 0.83 & 2.87 & 5.80 & 1.35 & 5.24 & 10.77 & 1.25 & 5.46 & 11.33 \\
 & $\text{SOMA}_{\text{ELoFTR}}$ & \bv{8.17} & \bv{20.30} & \bv{35.31} & \bv{5.20} & \bv{20.53} & \bv{39.97} & \bv{0.86} & \bv{2.96} & \bv{6.95} \\
\midrule

\multirow{2}{*}{Dense}
 & RoMa & 1.18 & 2.56 & 4.44 & 0.54 & 1.61 & 2.99 & 5.47 & 11.58 & 18.04 \\
 & $\text{MINIMA}_{\text{RoMa}}$ & 4.53 & 8.89 & 13.84 & 2.36 & 7.39 & 13.56 & 8.04 & 16.27 & 24.30 \\
\bottomrule
\end{tabular}

\vspace{1ex}
\parbox{\textwidth}{
    \footnotesize
    \textbf{Note:} Red bold text indicates the best results for each metric; blue text indicates that models fine-tuned on SOMA-0.1M (designated as $\text{SOMA}_{\text{Model}}$) outperform their corresponding original pre-trained versions; metrics are reported as percentages.
}
\end{table*}

\subsubsection{Evaluation Protocol}

\textbf{Ground Truth Calculation:} Since the images in the SOMA-Test and OSdataset are intrinsically aligned, we apply a random homography transformation to the left image ($I_L$) of each pair while maintaining the right image ($I_R$) as fixed to simulate realistic geometric distortions. To ensure consistent input data across all comparative experiments, we utilize fixed random seeds derived from the sample IDs during traversal to generate the transformation parameters; the specific settings for these perturbation parameters are as follows: Rotation follows a random angle within the range of $\left[-{108}^\circ,{108}^\circ\right]$ (i.e., $\pm 30\%$ of the full circle); Translation involves a random shift in both $x$ and $y$ directions within $\pm 30\%$ of the image width and height; Scale utilizes a random scaling factor between 0.7 and 1.3. Let $H_{dist}$ be the transformation matrix used to generate the perturbation. The ground truth homography matrix $H_{gt}$ from the transformed left image to the right image is then defined as the inverse of the perturbation matrix:
\begin{equation}
    H_{gt} = H_{pert}^{-1}
\end{equation}

\textbf{Evaluation Metrics:} During the evaluation phase, we employ RANSAC as a robust estimator to solve for the homography matrix, with parameters set to a reprojection error threshold of 1.5 pixels, a maximum of 10,000 iterations, and a confidence level of 0.9999. Following the standard protocol of MapGlue, we adopt the area under the curve (AUC) of the corner error as the quantitative evaluation metric. Specifically, we report the AUC values at error thresholds of 5 pixels, 10 pixels, and 20 pixels, which are denoted as AUC@5, AUC@10, and AUC@20, respectively.

\subsubsection{Baseline and Implementation Details}

To evaluate the contribution of the dataset, we selected current mainstream feature matching algorithms as baselines, covering three categories: sparse matching, semi-dense matching, and dense matching. For sparse matching, we selected two of the latest traditional multimodal matching methods, MSG and GLIFT. Regarding deep learning methods, we included representative models such as LightGlue, DiffGlue, and LightGlueStick, alongside methods specifically designed for multimodal remote sensing imagery, including ReDFeat and MapGlue. In addition, MambaGlue and $\text{MINIMA}_{\text{LG}}$ were incorporated, where $\text{MINIMA}_{\text{LG}}$ corresponds to a version of LightGlue retrained on other multimodal datasets. For consistency across sparse matching methods, the maximum number of extracted keypoints was fixed at 2,048. For semi-dense matching, we chose the transformer-based ELoFTR and XoFTR, as well as their retrained versions on other multimodal datasets, namely $\text{MINIMA}_{\text{ELoFTR}}$ and $\text{MINIMA}_{\text{XoFTR}}$. For dense matching, we selected RoMa and its retrained version, $\text{MINIMA}_{\text{RoMa}}$.

Training Settings and Comparative Strategy. To assess the impact of SOMA-1M on geometric feature learning, we adopted a comparative experimental design of pre-training versus SOMA fine-tuning. Traditional methods, RoMa, and the MINIMA series models were tested directly using their publicly available official weights as performance references without retraining. All other models were trained on the SOMA-0.1M dataset by loading their official pre-trained weights based on MegaDepth and then continuing training on SOMA-0.1M. All models trained on SOMA-0.1M utilized default parameter settings. During the data augmentation phase, the inputs for all models were resized to 512 $\times$ 512 pixels. Following the approach of LightGlue, we employed an online augmentation strategy to adapt to the arbitrary imaging directions of remote sensing imagery, which controlled the intensity of homography transformations by setting a difficulty coefficient of 0.7 and a translation scale of 1.0. Specifically, we incorporated $\pm 180^{\circ}$ omnidirectional rotation augmentation to force the models to learn rotation-invariant geometric features. Performance differences observed before and after fine-tuning on SOMA-0.1M are used to quantify the contribution of the dataset.

\begin{figure*}[!t] 
    \centering
    \includegraphics[width=1.00\textwidth]{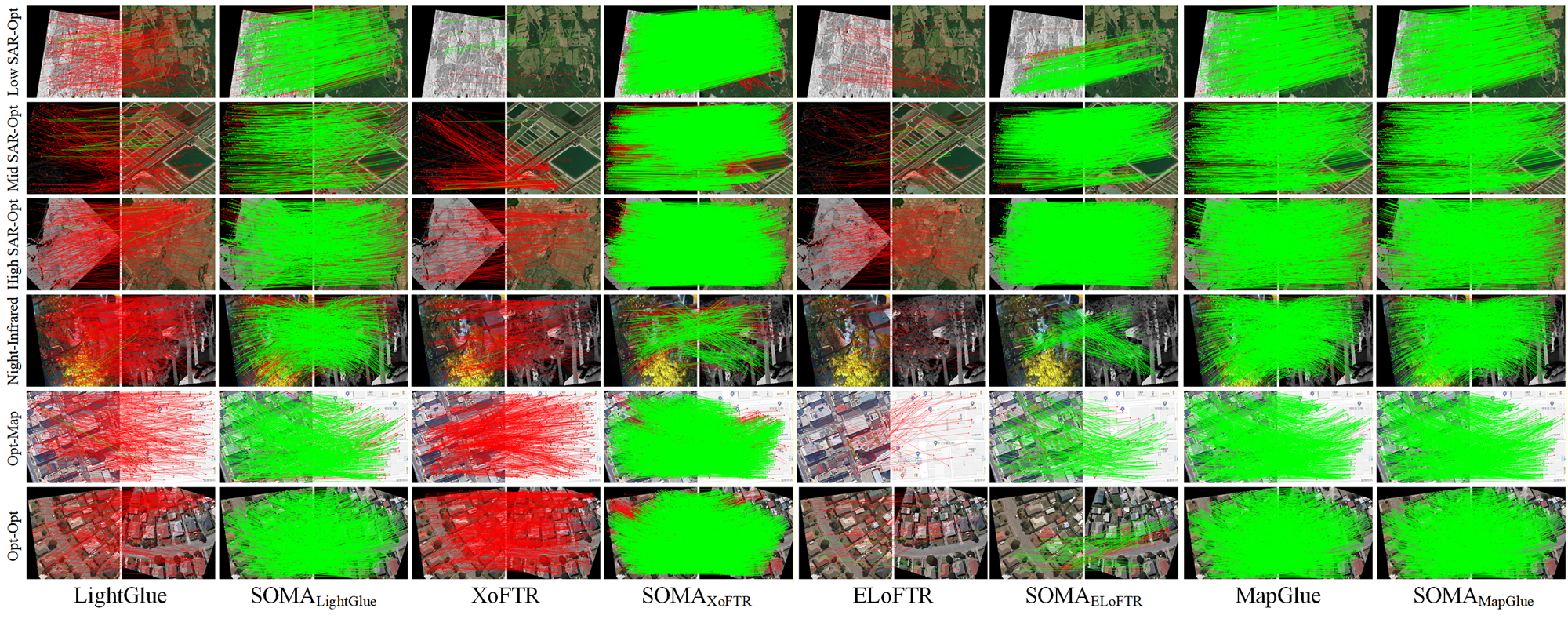}
    \caption{Qualitative comparison results of image matching.}
    \label{fig:matching qualitative comparison}
\end{figure*}

\subsubsection{Comparison Experiments}

As shown in Table \ref{tab:matching_quantitative_eval}, we conducted a quantitative evaluation of different matching algorithms across three test sets, including SOMA-Test, OSdataset, and SRIF. Red bold values represent the best performance for each metric, while blue values indicate that the fine-tuned models outperform their original pre-trained counterparts. The results show that models fine-tuned on the SOMA-0.1M dataset consistently achieve higher performance across all evaluation metrics. This addresses the limitations of models such as LightGlue, MambaGlue, XoFTR, and ELoFTR, which are primarily trained on natural scene datasets (e.g., MegaDepth) and therefore exhibit reduced robustness when applied to SAR–optical image pairs with strong radiometric differences and geometric distortions. After fine-tuning on SOMA-0.1M, the matching performance of these models on SAR–optical data improves substantially, exceeding that of traditional multimodal matching methods such as MSG and GLIFT, as well as the MINIMA series models trained on other multimodal datasets. Specifically, $\text{SOMA}_{\text{LightGlue}}$, $\text{SOMA}_{\text{MambaGlue}}$, $\text{SOMA}_{\text{XoFTR}}$, and $\text{SOMA}_{\text{ELoFTR}}$ achieve AUC@5px improvements of 5.9\%, 4.6\%, 4.6\%, and 7.88\%, respectively, on the SOMA-Test dataset, while their AUC@20px values increase by 31.04\%, 16.58\%, 22.06\%, and 33.70\%. Notably, both $\text{SOMA}_{\text{XoFTR}}$ and $\text{SOMA}_{\text{ELoFTR}}$ exhibit significant improvements after fine-tuning, indicating that the SOMA-0.1M dataset also substantially enhances the performance of semi-dense matching. $\text{SOMA}_{\text{MapGlue}}$ achieves the highest overall performance among all evaluated methods, reflecting the advantage of its architecture tailored for multimodal remote sensing data and its pre-training on map–optical imagery. Further fine-tuning on SOMA-0.1M leads to additional performance improvements, with AUC@5px, AUC@10px, and AUC@20px reaching 18.63\%, 37.25\%, and 55.42\%, respectively, surpassing the dense matching method RoMa and $\text{MINIMA}_{\text{RoMa}}$. In terms of cross-dataset generalization, all SOMA-series models show performance gains on the OSdataset. Notably, $\text{SOMA}_{\text{ELoFTR}}$ demonstrates the most significant improvements, increasing AUC@5px, AUC@10px, and AUC@20px by 5.20\%, 20.15\%, and 38.52\%, respectively. In zero-shot evaluations on the SRIF dataset, most models also demonstrate improved performance, except for $\text{SOMA}_{\text{XoFTR}}$. This exception is mainly attributed to the fact that the infrared–optical and depth–optical subsets in SRIF consist primarily of natural scenes, for which the original XoFTR model is already well optimized, thereby limiting the relative benefit of SOMA-based fine-tuning. Overall, the SOMA-0.1M dataset improves matching accuracy for SAR–optical imagery and enhances model robustness across multiple multimodal settings, including infrared–optical, map–optical, depth–optical, and day–night image pairs.

Fig. \ref{fig:matching qualitative comparison} presents a qualitative comparison between the SOMA-enhanced models and their original versions. Except for MapGlue, which already exhibits strong generalization ability, the SOMA-enhanced versions of the remaining models produce more accurate and stable matching results than their original counterparts, illustrating the benefit of SOMA-0.1M for multimodal remote sensing image matching.

\subsection{Image Fusion}

\begin{table*}[t]
\centering
\caption{Quantitative evaluation results of different fusion algorithms on SOMA-Test multi-resolution subsets and the OSdataset.}
\label{tab:fusion_quantitative_eval}

\footnotesize 
\setlength{\tabcolsep}{2.8pt} 
\renewcommand{\arraystretch}{1.2}

\begin{tabular}{l|cccc|cccc|cccc|cccc}
\toprule
\multirow{2}{*}{Method} & \multicolumn{4}{c|}{SOMA-Test-Low} & \multicolumn{4}{c|}{SOMA-Test-Mid} & \multicolumn{4}{c|}{SOMA-Test-High} & \multicolumn{4}{c}{OSdataset} \\
\cmidrule(lr){2-5} \cmidrule(lr){6-9} \cmidrule(lr){10-13} \cmidrule(lr){14-17}
 & FMI & MI & VIF & $Q^{AB/F}$ & FMI & MI & VIF & $Q^{AB/F}$ & FMI & MI & VIF & $Q^{AB/F}$ & FMI & MI & VIF & $Q^{AB/F}$ \\
\midrule
VSFF & 1.221 & 2.011 & 0.467 & 0.381 & 1.322 & 2.441 & 0.466 & 0.346 & 1.033 & 2.062 & 0.422 & 0.291 & 1.406 & 2.438 & 0.470 & 0.288 \\
BASHVS & 1.054 & 1.350 & 0.400 & 0.390 & \rb{2.103} & 2.493 & \rb{0.636} & 0.307 & 1.640 & 1.933 & \rb{0.554} & 0.312 & 1.919 & 2.373 & \rb{0.537} & 0.368 \\
RFN-Nest & 0.839 & 1.374 & 0.344 & 0.281 & 0.903 & 1.498 & 0.340 & 0.163 & 0.768 & 1.368 & 0.301 & 0.195 & 0.828 & 1.426 & 0.311 & 0.128 \\
$\text{SOMA}_{\text{RFN-Nest}}$ & \bv{0.852} & \bv{1.465} & \bv{0.378} & \bv{0.352} & \bv{1.567} & \bv{1.822} & \bv{0.447} & \bv{0.439} & \bv{1.099} & \bv{1.635} & \bv{0.389} & \bv{0.428} & \bv{1.703} & \bv{1.924} & \bv{0.478} & \bv{0.459} \\
LRRNet & 0.738 & 1.330 & 0.319 & 0.323 & 0.900 & 1.831 & 0.362 & 0.194 & 0.692 & 1.537 & 0.287 & 0.211 & 0.795 & 1.455 & 0.269 & 0.147 \\
$\text{SOMA}_{\text{LRRNet}}$ & \bv{1.039} & \bv{2.467} & \bv{0.538} & \bv{0.393} & \bv{1.676} & 1.554 & \bv{0.436} & \bv{0.503} & \bv{1.487} & \bv{1.958} & \bv{0.452} & \bv{0.502} & \bv{1.692} & \bv{1.824} & \bv{0.430} & \bv{0.542} \\
MUFusion & 0.576 & 1.175 & 0.311 & 0.246 & 0.784 & 0.959 & 0.276 & 0.117 & 0.617 & 1.026 & 0.256 & 0.140 & 0.850 & 1.116 & 0.285 & 0.130 \\
$\text{SOMA}_{\text{MUFusion}}$ & \bv{1.804} & \bv{2.206} & \bv{0.608} & \bv{0.403} & \bv{1.396} & \bv{1.810} & \bv{0.510} & \bv{0.245} & \bv{1.417} & \bv{2.035} & \bv{0.520} & \bv{0.254} & \bv{1.294} & \bv{1.761} & \bv{0.469} & \bv{0.233} \\
LENFusion & 1.288 & 3.052 & 0.594 & 0.470 & 1.511 & 2.659 & 0.483 & 0.432 & 1.234 & 2.633 & 0.466 & 0.432 & 1.631 & 2.574 & 0.469 & 0.508 \\
$\text{SOMA}_{\text{LENFusion}}$ & \bv{1.661} & \bv{3.604} & \bv{0.661} & \bv{0.486} & 1.335 & 2.438 & 0.477 & \bv{0.452} & 1.163 & 2.570 & 0.451 & 0.392 & 1.469 & 2.472 & \bv{0.479} & 0.486 \\
MMDRFuse & 1.356 & 2.399 & 0.577 & 0.505 & 1.619 & 2.097 & 0.523 & 0.518 & 1.361 & 2.030 & 0.494 & 0.478 & 1.692 & 1.889 & 0.483 & 0.549 \\
$\text{SOMA}_{\text{MMDRFuse}}$ & \bv{2.053} & \bv{3.567} & \rb{0.751} & \rb{0.531} & \bv{1.769} & 1.804 & 0.513 & \rb{0.564} & \bv{1.554} & 1.911 & \bv{0.507} & \rb{0.539} & \rb{1.975} & 1.832 & \bv{0.529} & \rb{0.606} \\
LUT-Fuse & 0.920 & 2.659 & 0.421 & 0.420 & 1.276 & 3.103 & 0.426 & 0.347 & 1.171 & 3.112 & 0.426 & 0.365 & 1.503 & 3.511 & 0.443 & 0.421 \\
$\text{SOMA}_{\text{LUT-Fuse}}$ & \rb{3.020} & \rb{5.273} & \bv{0.692} & \bv{0.515} & \bv{1.308} & \rb{3.250} & 0.357 & \bv{0.441} & \rb{1.695} & \rb{4.253} & 0.418 & \bv{0.455} & \bv{1.911} & \rb{3.966} & \bv{0.450} & \bv{0.541} \\
RPFNet & 0.775 & 1.378 & 0.340 & 0.161 & 1.155 & 1.636 & 0.355 & 0.108 & 0.875 & 1.469 & 0.316 & 0.108 & 1.228 & 1.600 & 0.354 & 0.111 \\
$\text{SOMA}_{\text{RPFNet}}$ & \bv{0.940} & \bv{1.554} & \bv{0.393} & \bv{0.358} & \bv{1.326} & \bv{2.158} & \bv{0.503} & \bv{0.312} & \bv{1.120} & \bv{1.806} & \bv{0.460} & \bv{0.291} & \bv{1.280} & \bv{1.911} & \bv{0.447} & \bv{0.321} \\
\bottomrule
\end{tabular}

\vspace{1ex}
\parbox{\textwidth}{
    \footnotesize 
    \textbf{Note:} Red bold text indicates the best results for each metric; blue text indicates that models fine-tuned on SOMA-0.1M (designated as $\text{SOMA}_{\text{Model}}$) outperform their corresponding original pre-trained versions.
}
\end{table*}

In this section, we evaluate SAR–optical image fusion performance using the SOMA-1M dataset. The experiments focus on assessing how effectively different models combine geometric information from SAR imagery with spectral information from optical imagery. To examine model behavior across different resolutions and sensor domains, we adopt a cross-resolution training and testing protocol. Specifically, all models are trained using only the low-resolution subset of SOMA-0.1M, which contains 33,334 image pairs (denoted as SOMA-0.1M-Low), and are then directly evaluated on the multi-resolution SOMA-Test set and the OSdataset. This experimental setting allows us to assess the robustness of the learned fusion representations when applied to data with different spatial resolutions and sensor characteristics.

\subsubsection{Evaluation Protocol}

Quantitative evaluation of SAR–optical image fusion is typically conducted without reference ground truth. For this reason, and due to the strong influence of speckle noise in SAR imagery, we do not adopt pixel-wise metrics such as PSNR, nor entropy-based metrics such as EN, which are sensitive to random noise. Instead, four commonly used non-reference fusion metrics are employed: feature mutual information (FMI), mutual information (MI), visual information fidelity (VIF), and gradient-based fusion quality ($Q^{AB/F}$). Larger values of these metrics indicate that the fused image preserves more information from the source modalities.

\subsubsection{Baseline and Implementation Details}

We select a set of representative image fusion methods as baselines, covering both traditional approaches and deep learning-based models. The traditional methods include VSFF and BASHVS. The deep learning baselines consist of RFN-Nest, LRRNet, MUFusion, LENFusion, MMDRFuse, LUT-Fusion, and RPFNet. Most of the deep learning models were originally developed for infrared–optical fusion and are provided with pre-trained weights obtained from natural scene datasets. In our experiments, these models are adapted to SAR–optical fusion without architectural modification. For training, the officially released pre-trained weights are loaded, followed by fine-tuning or continued training on the SOMA-0.1M-Low dataset. All training procedures follow the default hyperparameter settings reported in the original papers to ensure a fair comparison across different methods.

\begin{figure*}[!t] 
    \centering
    \includegraphics[width=1.00\textwidth]{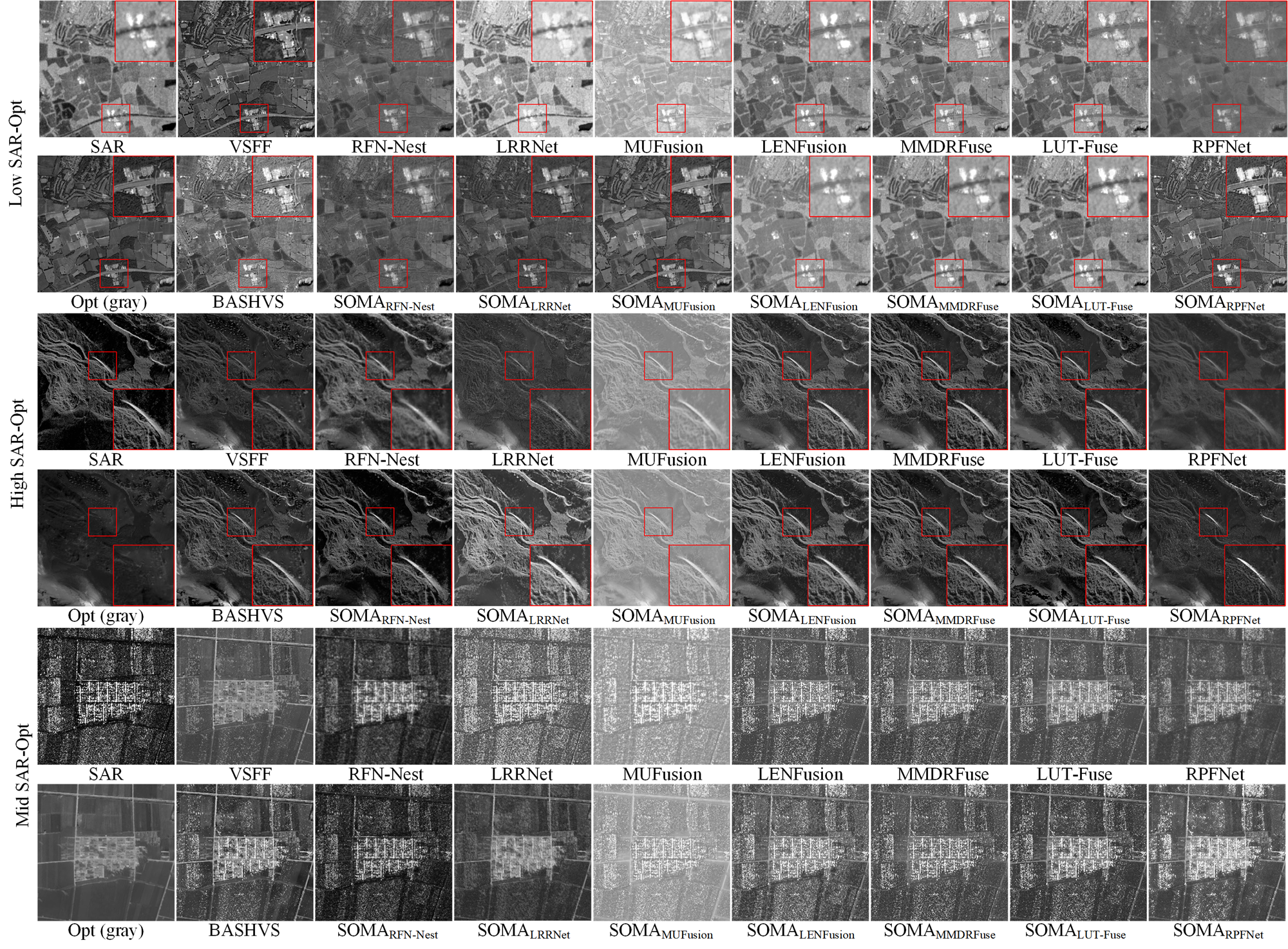}
    \caption{Qualitative comparison results of image fusion.}
    \label{fig:fusion qualitative comparison}
\end{figure*}

\subsubsection{Comparison Experiments}

Table \ref{tab:fusion_quantitative_eval} presents the quantitative evaluation results of various fusion algorithms across four test sets, including the SOMA-Test multi-resolution subsets (Low, Mid, and High) and the OSdataset. The results show that models fine-tuned on the SOMA-0.1M-Low dataset consistently achieve improved performance across all evaluation metrics. Particularly on the SOMA-Test-Low dataset, all SOMA-series models outperform their original counterparts in four key metrics: FMI, MI, VIF, and $Q^{AB/F}$. Among these, $\text{SOMA}_{\text{LUT-Fuse}}$ exhibits the most notable performance improvement, with increases of 228.26\%, 98.31\%, 64.37\%, and 22.62\% in FMI, MI, VIF, and $Q^{AB/F}$, respectively. In addition, it achieves the highest values among all compared methods in information-related metrics (FMI and MI). Meanwhile, $\text{SOMA}_{\text{MMDRFuse}}$ performs best in visual and structural metrics (VIF and $Q^{AB/F}$), with relative improvements of 30.16\% and 5.15\% over the original model. These observations demonstrate that the SOMA dataset effectively enhances pixel-level feature learning and multimodal information complementarity. Furthermore, when generalizing the models to the SOMA-Test-Mid, SOMA-Test-High, and OSdataset test sets, most fine-tuned models continue to outperform their original versions, indicating that training on low-resolution data enables the models to learn transferable and resolution-invariant complementary representations between SAR and optical imagery. This suggests a certain degree of cross-resolution generalization capability. Notably, on the middle- and high-resolution test sets, the traditional method BASHVS outperforms deep learning-based models in terms of the VIF metric. This result highlights a limitation of existing deep fusion models, which are primarily designed for infrared–optical fusion, when applied to SAR–optical scenarios. In particular, unsupervised training paradigms still face challenges in effectively handling strong SAR speckle noise.

Fig. \ref{fig:fusion qualitative comparison} presents a qualitative comparison between the SOMA-enhanced models and their original versions. After fine-tuning on the SOMA-0.1M-Low dataset, most models exhibit improved spatial detail in the fusion results, allowing SAR geometric structural information to be injected into optical imagery while maintaining good visual quality. This improvement can be clearly observed in models such as $\text{SOMA}_{\text{LRRNet}}$, $\text{SOMA}_{\text{MUFusion}}$, and $\text{SOMA}_{\text{RPFNet}}$. Among all methods, $\text{SOMA}_{\text{RPFNet}}$ shows the most visually consistent results, as it achieves a relatively balanced integration of SAR scattering characteristics and optical texture information, while preserving object boundaries. However, when applied to higher-resolution datasets, the suppression of large-scale speckle noise remains limited. As shown in the third group of experimental results, noticeable noise artifacts persist in the fused images. These observations indicate that current unsupervised fusion frameworks still face difficulties in simultaneously decoupling noise and preserving fine structural details. Without explicit constraints, current models struggle to suppress SAR-specific speckle noise while preserving subtle structural details, particularly at higher spatial resolutions.

\begin{figure}
    \centering
    \includegraphics[width=1.00\columnwidth]{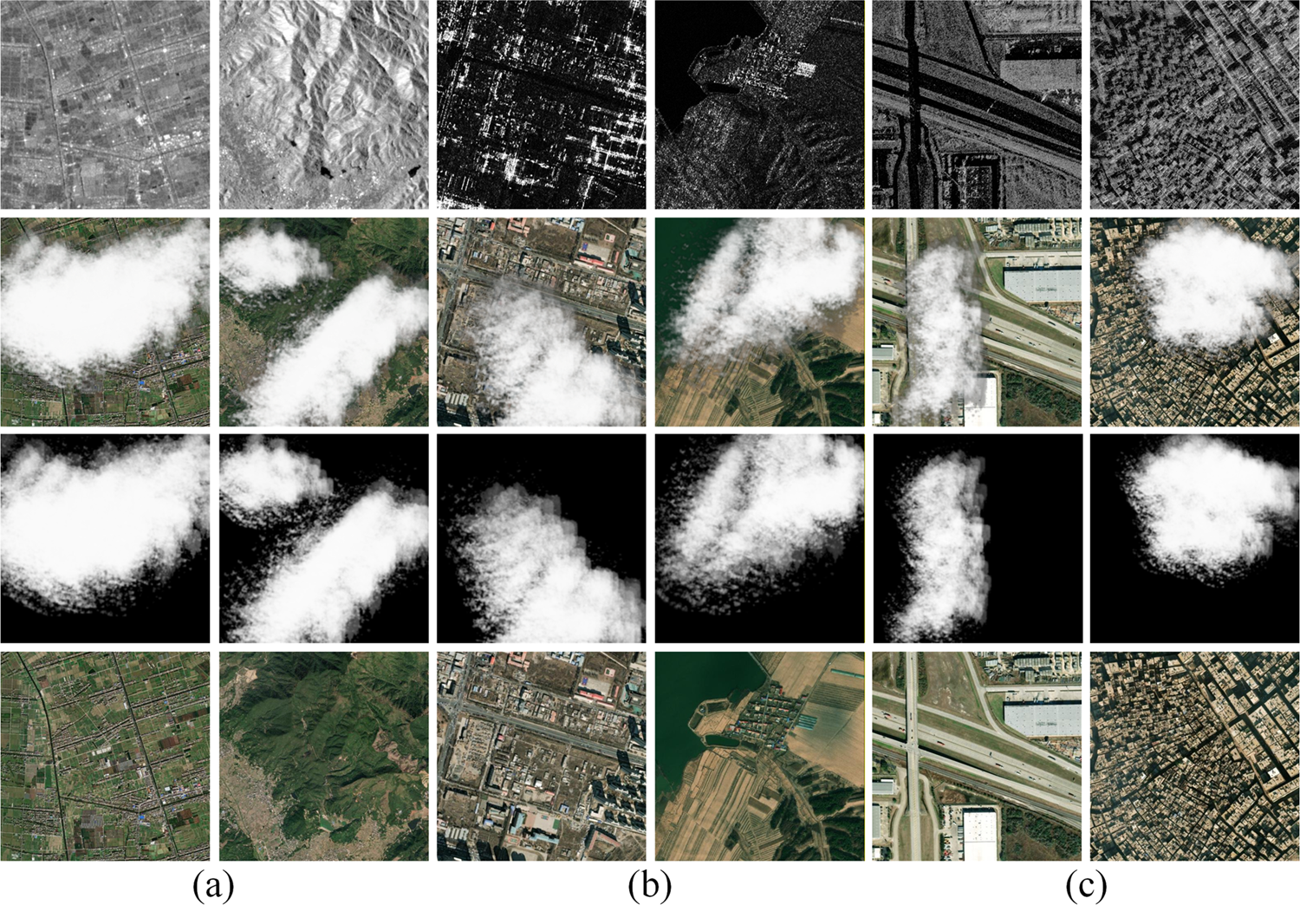}
    \caption{Low, mid, and high resolution cloud image pairs in the SOMA-0.1M dataset. (a) Low resolution, (b) Middle resolution, (c) High resolution.}
    \label{fig:cloud image}
\end{figure}

\subsection{SAR-Assisted Cloud Removal}

In this section, we evaluate the role of the SOMA-1M dataset in SAR-assisted optical cloud removal. Cloud coverage in optical imagery often leads to missing or severely degraded surface information, whereas SAR imagery remains observable under cloudy conditions. Based on this property, we design a multimodal cloud removal experiment in which SAR data are used as an additional modality to support the reconstruction of cloud-covered optical regions. To construct paired training data at scale, we adopt a synthetic data generation scheme. Specifically, simulated clouds with varying densities and spatial patterns are overlaid onto 100,000 cloud-free optical images from SOMA-0.1M. This process produces paired samples consisting of SAR images, cloudy optical images, corresponding cloud masks, and cloud-free optical references, as illustrated in Fig. \ref{fig:cloud image}. The resulting dataset provides a controlled benchmark for evaluating multimodal cloud removal methods.

\subsubsection{Evaluation Metrics}

Since the experiments are conducted on synthetic data with corresponding cloud-free optical images available as ground truth, we adopt four quantitative metrics to evaluate reconstruction performance. PSNR and SSIM are used to assess the similarity between the reconstructed images and the ground truth at the pixel and structural levels. Spectral Angle Mapper (SAM) measures spectral distortion by computing the angular difference between reconstructed and reference pixel vectors in color space. In addition, Mean Absolute Error (MAE) is used to quantify pixel-wise reconstruction error. For PSNR and SSIM, higher values indicate better performance, while lower values of SAM and MAE correspond to more accurate reconstruction.

\subsubsection{Baseline and Implementation Details}

We select six commonly used deep learning methods for optical cloud removal as baselines, including CNN-based models (Dsen-CR, GLF-CR, Align-CR, ACA-CRNet, and HPN-CR) and the diffusion-based model EMRDM. Most of these methods were originally developed for 13-band Sentinel-2 multispectral data from the SEN12MS-CR dataset. To adapt them to the 3-band RGB format of SOMA-0.1M, the number of input channels in the first layer of each network is modified accordingly. All other network configurations follow the original implementations. During training, images are randomly cropped into patches of size 256 $\times$ 256. During inference, an overlapping sliding-window strategy is employed to reduce boundary artifacts. The window size is set to 256 $\times$ 256 with a stride of 224, resulting in a 32-pixel overlap between adjacent patches. Predictions in overlapping regions are fused using weighted averaging to ensure spatial continuity in the reconstructed images.

To evaluate the effect of SAR information, we conduct a comparative experiment between single-modal optical cloud removal and SAR-assisted cloud removal. For the single-modal setting, the SAR input branch is removed or replaced with zero-padding, while keeping the remaining network architecture and parameters unchanged. This design ensures a fair comparison between the two input configurations.

\subsubsection{Comparison Experiments}

\begin{figure*}[!t] 
    \centering
    \includegraphics[width=1.00\textwidth]{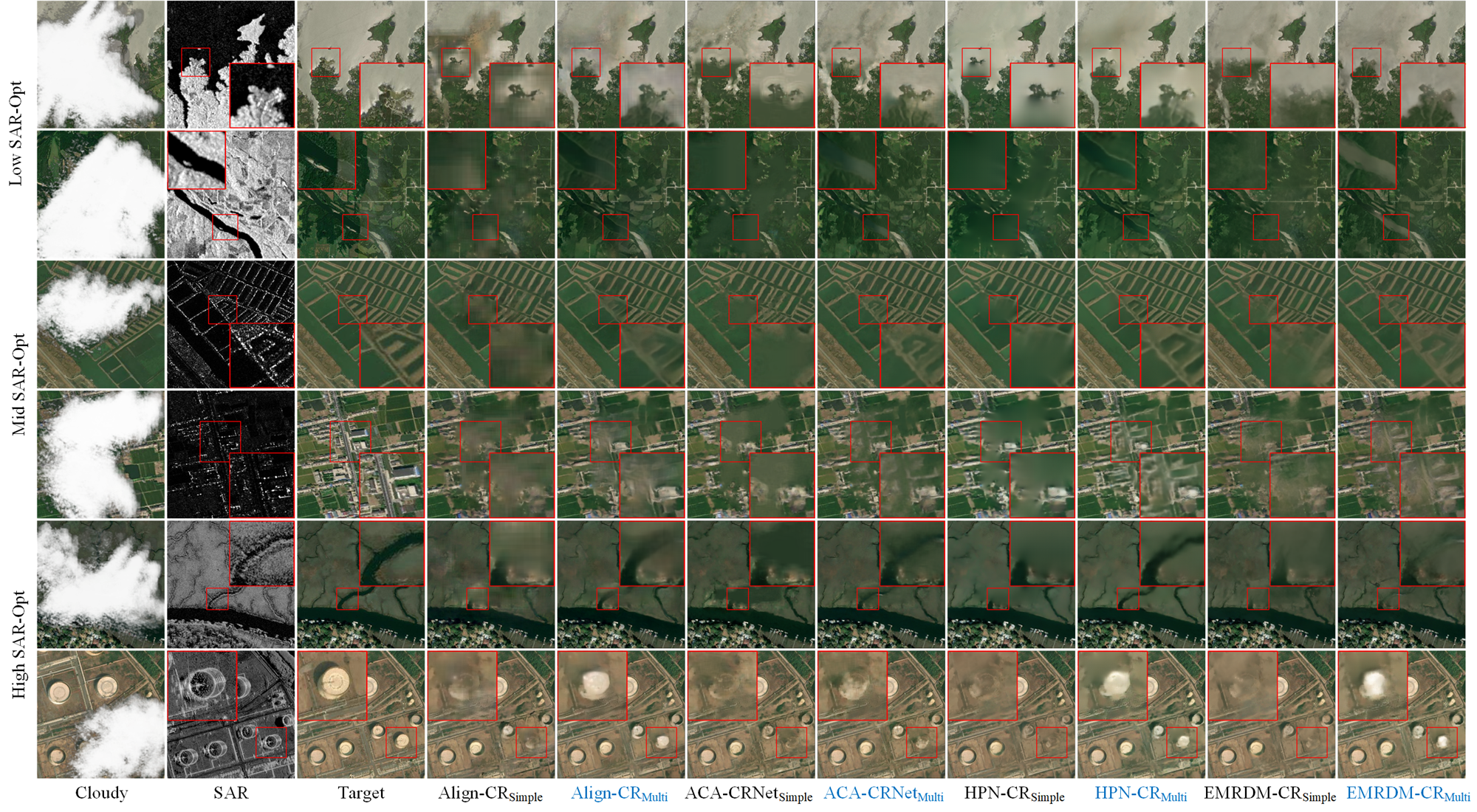}
    \caption{Qualitative comparison results of cloud removal.}
    \label{fig:cloud_removal}
\end{figure*}

\begin{table}[t]
\centering
\caption{Quantitative comparison results of single-modal and SAR-assisted cloud removal for various algorithms.}
\label{tab:cloud_removal}

\footnotesize 
\setlength{\tabcolsep}{5pt} 
\renewcommand{\arraystretch}{1.2}

\begin{tabular}{lcccc}
\toprule
\multicolumn{1}{c}{Method} & PSNR $\uparrow$ & SSIM $\uparrow$ & MAE $\downarrow$ & SAM $\downarrow$ \\
\midrule
$\text{Dsen-CR}_{\text{Simple}}$ & 24.921 & 0.858 & 0.028 & 2.242 \\
$\text{Dsen-CR}_{\text{Multi}}$ & \bv{25.246} & \bv{0.867} & 0.028 & 2.272 \\
\midrule
$\text{GLF-CR}_{\text{Simple}}$ & 23.567 & 0.809 & 0.036 & 2.503 \\
$\text{GLF-CR}_{\text{Multi}}$ & \bv{25.003} & \bv{0.858} & \bv{0.027} & \bv{2.296} \\
\midrule
$\text{Align-CR}_{\text{Simple}}$ & 25.358 & 0.867 & 0.026 & 2.041 \\
$\text{Align-CR}_{\text{Multi}}$ & \bv{25.951} & \bv{0.874} & \rb{0.023} & \bv{1.998} \\
\midrule
$\text{ACA-CRNet}_{\text{Simple}}$ & 25.830 & 0.878 & 0.024 & 1.993 \\
$\text{ACA-CRNet}_{\text{Multi}}$ & \bv{26.283} & \rb{0.885} & \bv{0.022} & \bv{1.911} \\
\midrule
$\text{HPN-CR}_{\text{Simple}}$ & 28.739 & 0.884 & 0.025 & 1.828 \\
$\text{HPN-CR}_{\text{Multi}}$ & \rb{28.849} & \rb{0.885} & \bv{0.024} & \bv{1.626} \\
\midrule
$\text{EMRDM}_{\text{Simple}}$ & 26.197 & 0.867 & 0.026 & 1.371 \\
$\text{EMRDM}_{\text{Multi}}$ & \bv{26.800} & \bv{0.871} & \bv{0.024} & \rb{1.267} \\
\bottomrule
\end{tabular}

\vspace{1ex}
\footnotesize 
\parbox{\linewidth}{
    \small
    \textbf{Note:} Red bold text indicates the best results for each metric; blue text indicates that SAR-assisted cloud removal models ($\text{Model}_{\text{Multi}}$) outperform their single-modal optical counterparts ($\text{Model}_{\text{Simple}}$).
}
\end{table}

As shown in Table \ref{tab:cloud_removal}, we report quantitative results of different cloud removal methods evaluated on the SOMA-Test dataset. Methods incorporating SAR information consistently outperform their single-modal counterparts across all metrics, indicating that the additional SAR modality provides useful complementary information for cloud-covered regions. Among the evaluated methods, GLF-CR exhibits the largest performance improvement when SAR data are introduced. Specifically, $\text{GLF-CR}_{\text{Multi}}$ achieves improvements of 6.09\% in PSNR, 6.06\% in SSIM, and reductions of 25.00\% and 8.27\% in MAE and SAM, respectively, compared with $\text{GLF-CR}_{\text{Simple}}$. This result suggests that accurate SAR–optical alignment in SOMA-0.1M is beneficial for exploiting cross-modal information during cloud removal. Differences across model architectures can also be observed. $\text{HPN-CR}_{\text{Multi}}$ achieves the highest PSNR and SSIM values, while the diffusion-based $\text{EMRDM}_{\text{Multi}}$ performs best in terms of SAM. This indicates that CNN-based and diffusion-based models emphasize different aspects of the reconstruction task. CNN-based methods tend to better preserve local structures and pixel-level fidelity, which is reflected in PSNR and SSIM, whereas diffusion-based models show advantages in maintaining spectral consistency, as indicated by lower SAM values.

Fig. \ref{fig:cloud_removal} presents qualitative comparisons between single-modal and SAR-assisted cloud removal. Compared with optical-only methods, SAR-assisted approaches better preserve the geometric structure and texture continuity in regions affected by dense cloud coverage.

\begin{table*}[t]
\centering
\caption{Quantitative comparison of unsupervised and supervised SAR-to-optical translation.}
\label{tab:translation_eval}

\footnotesize
\setlength{\tabcolsep}{2pt} 
\renewcommand{\arraystretch}{1.8}

\begin{tabular}{c|c|ccccc|ccccc|ccccc}
\toprule
\multirow{2}{*}{\rotatebox{90}{Category}} & \multirow{2}{*}{Method} & \multicolumn{5}{c|}{SOMA-Test-Low} & \multicolumn{5}{c|}{SOMA-Test-Mid} & \multicolumn{5}{c}{SOMA-Test-High} \\
\cmidrule{3-17}
 & & PSNR$\uparrow$ & SSIM$\uparrow$ & SAM$\downarrow$ & LPIPS$\downarrow$ & FID$\downarrow$ & PSNR$\uparrow$ & SSIM$\uparrow$ & SAM$\downarrow$ & LPIPS$\downarrow$ & FID$\downarrow$ & PSNR$\uparrow$ & SSIM$\uparrow$ & SAM$\downarrow$ & LPIPS$\downarrow$ & FID$\downarrow$ \\
\midrule

\multirow{3}{*}{\rotatebox{90}{Unsupervised}} 
 & CycleGAN & 14.075 & 0.144 & 9.119 & 0.514 & 59.160 & 13.959 & 0.226 & 7.686 & 0.438 & 76.349 & 12.171 & 0.175 & 9.601 & 0.502 & 65.259 \\
 & CUT & 14.220 & 0.155 & 8.919 & 0.512 & \rb{53.621} & 14.169 & 0.226 & 7.887 & 0.443 & 69.747 & 11.995 & 0.159 & 9.535 & 0.531 & 64.999 \\
 & UNSB & 14.023 & 0.155 & 8.116 & 0.507 & 62.376 & 13.666 & 0.206 & 8.209 & 0.456 & 81.090 & 12.802 & 0.196 & 9.090 & 0.520 & 66.302 \\
\midrule

\multirow{3}{*}{\rotatebox{90}{Supervised}} 
 & Pix2pix & 17.378 & 0.283 & 6.804 & 0.458 & 68.223 & 15.857 & 0.299 & 7.277 & 0.415 & 97.451 & 15.763 & 0.258 & 7.232 & 0.464 & 90.401 \\
 & Pix2pixHD & \rb{17.860} & 0.320 & \rb{6.140} & \rb{0.417} & 58.274 & \rb{16.911} & 0.361 & \rb{6.305} & \rb{0.361} & \rb{67.444} & \rb{16.418} & 0.301 & \rb{6.327} & \rb{0.394} & \rb{63.428} \\
 & BBDM & 17.416 & \rb{0.350} & 6.765 & 0.499 & 75.504 & 15.718 & \rb{0.372} & 8.339 & 0.451 & 84.292 & 15.278 & \rb{0.315} & 7.498 & 0.505 & 75.778 \\
\bottomrule
\end{tabular}

\vspace{1ex}
\footnotesize 
\parbox{\linewidth}{
    \textbf{Note:} Red bold text indicates the best results for each metric.
}
\end{table*}

\subsection{SAR-to-Optical Translation}

This section evaluates the generative capabilities of SOMA-1M in cross-modal mapping. SAR-to-optical translation aims to convert complex SAR scattering features into interpretable optical textures aligned with human visual perception. To demonstrate the advantage of high-quality paired data over traditional unpaired datasets, we conducted a comparative study between unsupervised and supervised methods. By leveraging the pixel-level aligned data from SOMA-0.1M, we aim to show that while unsupervised methods reduce data requirements, supervised training on paired data yields superior translation results, characterized by both realistic optical textures and precise geographic structural consistency.

\subsubsection{Evaluation Metrics}

To quantitatively assess the generation results, we evaluate performance across two dimensions: perceptual quality and physical fidelity. In the perceptual dimension, we employ FID and LPIPS to measure the realism of deep feature distributions and visual similarity, respectively. In the physical dimension, SSIM and PSNR are utilized to assess the pixel-level and structural consistency between the generated content and the ground truth. Crucially, SSIM and PSNR serve as the primary indicators for verifying the advantages of supervised methods. Unsupervised methods, lacking the pixel-level alignment constraints provided by SOMA-0.1M, frequently suffer from geometric deformations or hallucinations, which are directly reflected in their inferior performance on these structural metrics.

\begin{figure*}[!t] 
    \centering
    \includegraphics[width=1.00\textwidth]{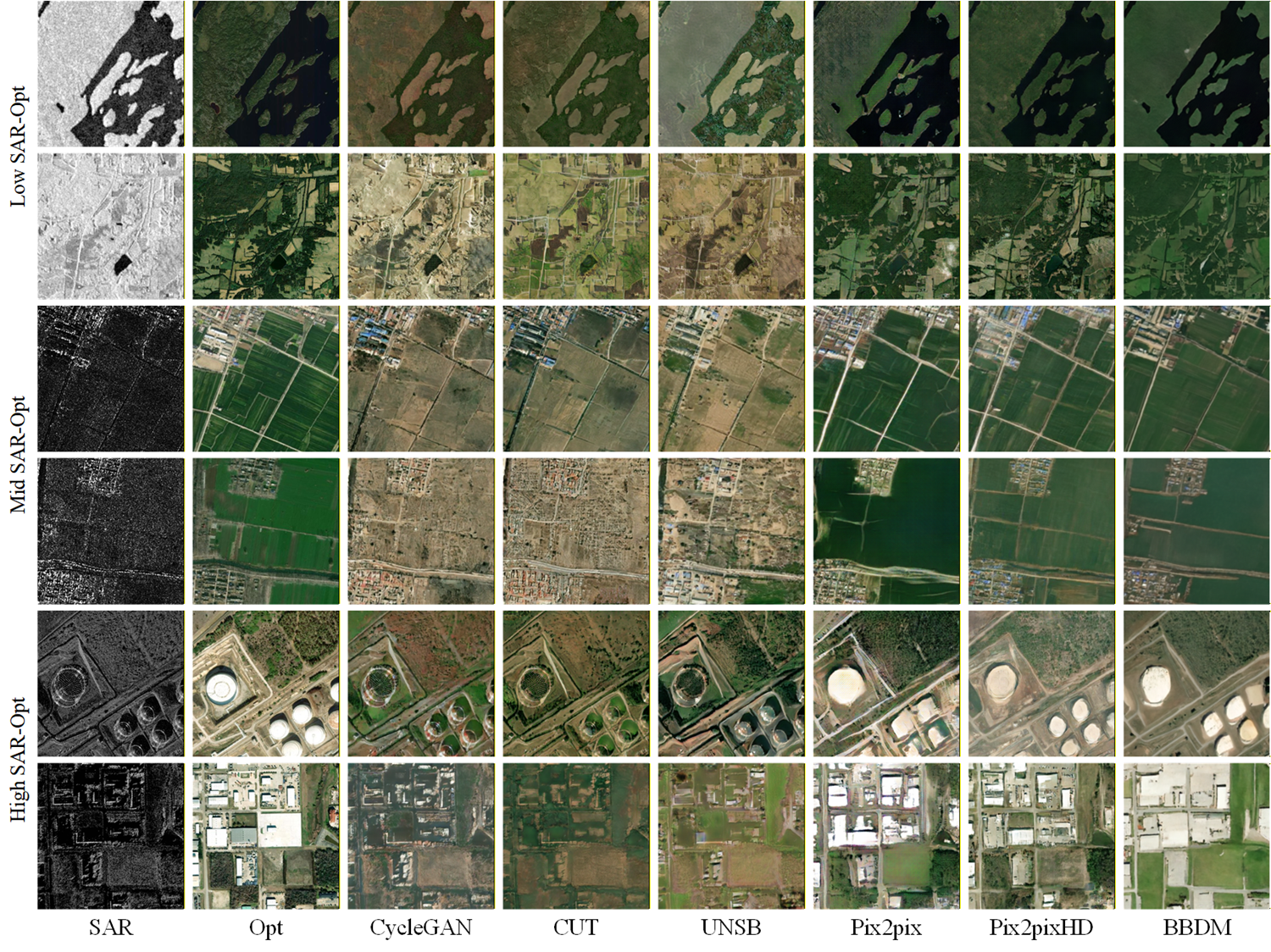}
    \caption{Qualitative comparison results of image translation.}
    \label{fig:image translation}
\end{figure*}

\subsubsection{Baseline and Implementation Details}

We selected six representative baselines, ranging from classic GANs to state-of-the-art diffusion models. These include unsupervised methods (CycleGAN, CUT, and UNSB) and supervised methods (pix2pix, pix2pixHD, and BBDM).

To account for the scale-dependent scattering characteristics of SAR imagery, we train separate models for the 0.5 m, 3 m, and 10 m subsets of SOMA-0.1M. High-resolution SAR images contain rich local structural variations caused by geometric scattering effects such as layover and double-bounce. In contrast, low-resolution images exhibit smoother intensity distributions dominated by large-scale spatial patterns. We find that a single model trained across these resolutions struggles to handle these heterogeneous characteristics simultaneously, often leading to degraded reconstruction quality at one or more scales. Training resolution-specific models therefore provides a more stable and effective solution. This observation supports the hierarchical multi-resolution design adopted in SOMA-1M.

\begin{table*}[t]
\centering
\caption{Quantitative evaluation results of image matching methods across different resolution subsets. AUC metrics are reported as percentages.}
\label{tab:matching_resolution}

\footnotesize
\setlength{\tabcolsep}{5pt} %
\renewcommand{\arraystretch}{1.2} %

\begin{tabular}{lccccccccc}
\toprule
\multirow{2}{*}{Method} & \multicolumn{3}{c}{SOMA-Test-Low} & \multicolumn{3}{c}{SOMA-Test-Mid} & \multicolumn{3}{c}{SOMA-Test-High} \\
\cmidrule(lr){2-4} \cmidrule(lr){5-7} \cmidrule(lr){8-10}
 & @5px & @10px & @20px & @5px & @10px & @20px & @5px & @10px & @20px \\
\midrule
$\text{SOMA}_{\text{ReDFeat}}$ & 2.49 & 5.75 & 9.94 & 0.33 & 0.72 & 1.45 & 0.94 & 2.68 & 5.30 \\
$\text{SOMA}_{\text{LightGlue}}$ & 12.91 & 31.37 & 49.67 & 1.85 & 8.99 & 23.05 & 4.84 & 15.38 & 29.71 \\
$\text{SOMA}_{\text{MambaGlue}}$ & 9.53 & 18.91 & 27.59 & 1.02 & 3.52 & 7.31 & 3.50 & 8.89 & 15.50 \\
$\text{SOMA}_{\text{DiffGlue}}$ & 8.48 & 19.89 & 31.33 & 0.75 & 3.48 & 9.22 & 2.39 & 7.30 & 15.54 \\
$\text{SOMA}_{\text{LightGlueStick}}$ & 11.36 & 28.11 & 45.70 & 1.07 & 4.83 & 14.01 & 2.78 & 9.71 & 21.40 \\
$\text{SOMA}_{\text{MapGlue}}$ & \textbf{34.68} & \textbf{55.58} & \textbf{69.99} & \textbf{8.19} & \textbf{24.98} & \textbf{45.45} & \textbf{13.13} & \textbf{31.28} & \textbf{50.91} \\
$\text{SOMA}_{\text{XoFTR}}$ & 12.16 & 23.81 & 34.52 & 1.84 & 9.62 & 22.83 & 3.63 & 12.45 & 23.99 \\
$\text{SOMA}_{\text{ELoFTR}}$ & 16.04 & 32.50 & 46.84 & 2.97 & 11.80 & 27.81 & 5.58 & 16.71 & 31.93 \\
\bottomrule
\end{tabular}
\end{table*}

\begin{table*}[t]
\centering
\caption{Quantitative evaluation results of SAR-assisted cloud removal methods across different resolution subsets.}
\label{tab:cloud_removal_resolution}

\footnotesize
\setlength{\tabcolsep}{3.5pt}
\renewcommand{\arraystretch}{1.2}

\begin{tabular}{lcccccccccccc}
\toprule
\multirow{2.5}{*}{Method} & \multicolumn{4}{c}{SOMA-Test-Low} & \multicolumn{4}{c}{SOMA-Test-Mid} & \multicolumn{4}{c}{SOMA-Test-High} \\
\cmidrule(lr){2-5} \cmidrule(lr){6-9} \cmidrule(lr){10-13}
 & PSNR$\uparrow$ & SSIM$\uparrow$ & MAE$\downarrow$ & SAM$\downarrow$ & PSNR$\uparrow$ & SSIM$\uparrow$ & MAE$\downarrow$ & SAM$\downarrow$ & PSNR$\uparrow$ & SSIM$\uparrow$ & MAE$\downarrow$ & SAM$\downarrow$ \\
\midrule
$\text{Dsen-CR}_{\text{Multi}}$ & 25.926 & 0.871 & 0.025 & 2.236 & 25.318 & 0.867 & 0.027 & 2.130 & 24.496 & 0.863 & 0.030 & 2.450 \\
$\text{GLF-CR}_{\text{Multi}}$ & 25.548 & 0.855 & 0.026 & 2.315 & 25.132 & 0.864 & 0.027 & 2.097 & 24.331 & 0.854 & 0.030 & 2.474 \\
$\text{Align-CR}_{\text{Multi}}$ & 26.486 & 0.870 & 0.022 & 2.003 & 26.091 & 0.881 & \textbf{0.022} & 1.829 & 25.277 & 0.871 & 0.026 & 2.163 \\
$\text{ACA-CRNet}_{\text{Multi}}$ & 26.929 & \textbf{0.886} & \textbf{0.020} & 1.881 & 26.280 & 0.886 & \textbf{0.022} & 1.779 & 25.641 & \textbf{0.883} & \textbf{0.024} & 2.072 \\
$\text{HPN-CR}_{\text{Multi}}$ & \textbf{29.184} & 0.881 & 0.023 & 1.664 & \textbf{29.161} & \textbf{0.892} & 0.023 & 1.351 & \textbf{28.203} & 0.882 & 0.027 & 1.862 \\
$\text{EMRDM}_{\text{Multi}}$ & 27.319 & 0.866 & 0.022 & \textbf{1.243} & 27.022 & 0.879 & 0.024 & \textbf{1.143} & 26.059 & 0.867 & 0.027 & \textbf{1.414} \\
\bottomrule
\end{tabular}
\end{table*}

\subsubsection{Comparison Experiments}

As shown in Table \ref{tab:translation_eval}, supervised methods trained on SOMA-0.1M consistently achieve higher quantitative performance than unsupervised approaches in the SAR-to-optical translation task. Across different methods, pix2pix, pix2pixHD, and BBDM outperform CycleGAN, CUT, and UNSB in terms of SSIM and PSNR, indicating better preservation of structural consistency between SAR and optical images. This performance gap suggests that pixel-level aligned supervision plays an important role in constraining the geometric correspondence during generation. In terms of perceptual quality, pix2pixHD achieves the lowest FID and LPIPS scores in the medium- and high-resolution subsets, which can be attributed to its multi-scale generator design. This architecture is more effective in modeling high-frequency details that are prominent in higher-resolution SAR imagery. When comparing results across different resolutions, the high-resolution subset generally exhibits higher FID scores than the low-resolution subset, reflecting the increased difficulty of modeling complex geometric scattering effects, such as building layover, at finer spatial scales. These observations indicate that a single-scale model is insufficient to handle the diverse scattering characteristics across resolutions, motivating the multi-resolution design adopted in SOMA-1M. Overall, supervised methods trained with high-precision aligned data achieve more reliable translation results in both structural fidelity and perceptual quality.

The qualitative results in Fig. \ref{fig:image translation} further illustrate the advantages of supervised methods trained with pixel-level aligned SAR–optical pairs from SOMA-0.1M. These methods better preserve the geometric structure of complex objects, such as rivers and buildings. In contrast, unsupervised methods often suffer from geometric distortions or spurious textures in medium- and high-resolution cases, where scattering effects are more pronounced. This comparison highlights the importance of high-quality paired data in mitigating speckle-related artifacts and improving the reconstruction of fine structural details.

\section{Discussion}

\subsection{Impact Analysis of Resolution on Hierarchical Tasks}

To investigate the impact of spatial resolution on multimodal downstream tasks, we supplemented the testing results of image matching and cloud removal methods across different resolution subsets, as shown in Table \ref{tab:matching_resolution} and Table \ref{tab:cloud_removal_resolution}. Combining results from Tables \ref{tab:fusion_quantitative_eval}, \ref{tab:translation_eval}, \ref{tab:matching_resolution}, and \ref{tab:cloud_removal_resolution}, the impact of resolution enhancement on geometric, pixel-level, and generative tasks exhibits a clear task-dependency. In the image matching task, performance significantly declines as resolution increases; for example, the AUC@20px of $\text{SOMA}_{\text{MapGlue}}$ is 69.99\% at low resolution but drops to 45.45\% and 50.91\% at middle and high resolutions, respectively. This indicates that the severe layover and speckle noise in high-resolution SAR imagery introduce complex geometric nonlinearities, hindering feature alignment. The image fusion task is similarly constrained by high-resolution noise. $\text{SOMA}_{\text{MMDRFuse}}$, which performs excellently in terms of visual fidelity, sees its VIF drop from 0.751 at low resolution to 0.507 at high resolution. Although the model captures cross-scale complementary features, the physical complexity of high-resolution scenes limits the ceiling of visual metrics.

In contrast, the cloud removal task proves to be the most robust, generally performing better on low-resolution datasets. Taking $\text{HPN-CR}_{\text{Multi}}$ as an example, its SSIM remains consistently between 0.88 and 0.89 across different resolutions, while the fluctuations in SAM are slightly larger, ranging from 1.35 to 1.86. This stability is attributed to the steady structural prior provided by the all-weather penetration of SAR imagery. Since cloud removal essentially relies on SAR structure to recover optical data, the inherent physical constraints of this cross-modal mapping limit the extent to which spatial resolution can improve spectral reconstruction accuracy, resulting in stable performance across scales. Conversely, image translation exhibits strong scale-dependency; the FID of Pix2pixHD fluctuates from 58.27 at low resolution to 67.44 at medium resolution, reflecting a clear degradation in generation quality during the scale transition. Aggregating the performance across all tasks, a sensitivity gradient emerges in the following order: $\text{Image Matching} > \text{Image Translation} > \text{Image Fusion} > \text{Cloud Removal}$. This multi-dimensional variance in sensitivity underscores the necessity of the multi-resolution hierarchical system in SOMA-1M, given the inherent limitation of single-scale models in simultaneously capturing fine-grained geometric details and broad-scale semantic context.

\subsection{Future Work}

While the SOMA-0.1M subset, representing less than 10\% of the total scale, has already demonstrated strong model activation capabilities across various downstream tasks, the full SOMA-1M dataset with its 1.3 million strictly aligned image pairs holds immense potential for future scientific exploration. Future research could utilize this massive volume of aligned samples for large-scale self-supervised pre-training. Specifically, techniques such as masked image modeling or contrastive learning can be employed to mine deep cross-modal universal representations, thereby addressing performance bottlenecks in matching and translation caused by incomplete feature decoupling in high-resolution scenarios. Simultaneously, by leveraging the absolute latitude and longitude coordinates preserved in the dataset, further exploration could be conducted on the deep coupling of geographic location encoding and cross-modal features. This effort aims to facilitate the construction of a remote sensing foundation model with global spatial perception capabilities, empowering macro-geospatial tasks such as temperature prediction, population density estimation, and global-scale change monitoring.

\section{Conclusion}

To mitigate the scarcity of large-scale, high-precision aligned data in remote sensing, we introduced SOMA-1M, a multi-resolution SAR-optical dataset with pixel-level alignment. Comprising over 1.3 million image pairs across 12 global land-cover categories, SOMA-1M sets a new benchmark for scene diversity and complexity. We validated the dataset's efficacy through comprehensive evaluations across four tasks: image matching, fusion, cloud removal, and translation. Notably, matching models trained on SOMA-1M achieved state-of-the-art performance, significantly outperforming existing methods.

Our analysis further reveals distinct sensitivity patterns regarding spatial resolution: low-level tasks (e.g., matching) are highly sensitive to micro-geometric distortions, whereas high-level semantic tasks (e.g., cloud removal) depend more on macroscopic structural fidelity. This insight validates the necessity of our multi-resolution hierarchical framework, as single-scale models fail to address the trade-off between fine-grained precision and global generalization. Beyond serving as a robust training resource, SOMA-1M provides reliable ground truth for emerging research in multimodal foundation models. Future work will focus on leveraging SOMA-1M for large-scale pre-training to advance intelligent perception in diverse observation scenarios.

\printcredits

\section*{Funding}

This work was supported in part by the Key Program of the National Key Research and Development Program of China Grant 2024YFB3909001, the Key Program of the National Natural Science Foundation of China under project number 42530106, the Program of the National Natural Science Foundation of China under Project 42401534 and Project 42471470, the LIESMARS Special Research Funding, the Key Laboratory of the Ministry of Education on Application of Artificial Intelligence in Equipment(2024-AAIEKF02-02), and the Key Laboratory of National Geographic Census and Monitoring, Ministry of Natural Resources (2025NGCM07).

\bibliographystyle{elsarticle-num} 
\bibliography{cas-refs.bib}

\end{document}